\documentclass[11pt, a4paper, logo, copyright]{googledeepmind}

\pdftrailerid{redacted}

\makeatletter
\providecommand\bibentry[1]{\nocite{#1}\fullcite{#1}}
\makeatother

\usepackage{kantlipsum, lipsum}
\usepackage{dsfont}
\usepackage{gdm-colors}
\usepackage[utf8]{inputenc}   
\usepackage{newunicodechar}   
\usepackage{amssymb}          
\usepackage{CJKutf8}
\usepackage{tcolorbox}
\tcbuselibrary{breakable}
\usepackage{enumitem}
\usepackage{arydshln}
\usepackage[flushleft]{threeparttable}
\usepackage{wasysym,marvosym}

\usepackage{svg}

\definecolor{catBlue}{HTML}{2A6FB5}
\definecolor{catPurple}{HTML}{6E3FA3}
\definecolor{catTeal}{HTML}{1F6F6B}
\definecolor{catRed}{HTML}{C0392B}
\definecolor{cellGreen}{HTML}{6CC180}
\definecolor{cellGray}{HTML}{ECECEC}

\usepackage{graphicx}

\usepackage{caption}
\usepackage{xcolor}
\usepackage{soul}

\soulregister\cite7
\soulregister\ref7
\soulregister\cref7

\usepackage[
  style=authoryear-comp,
  sorting=ynt,
  natbib=true,
  backend=bibtex,
  maxcitenames=2,
  maxbibnames=99,
  uniquelist=false,
  uniquename=false,
  giveninits=true,
  dashed=false
]{biblatex}
\let\cite\textcite

\DeclareFieldFormat{citehyperref}{%
  \DeclareFieldAlias{bibhyperref}{noformat}%
  \bibhyperref{#1}}
\DeclareFieldFormat{textcitehyperref}{%
  \DeclareFieldAlias{bibhyperref}{noformat}%
  \bibhyperref{%
    #1%
    \ifbool{cbx:parens}
      {\bibcloseparen\global\boolfalse{cbx:parens}}
      {}}}
\savebibmacro{cite}
\savebibmacro{textcite}
\renewbibmacro*{cite}{%
  \printtext[citehyperref]{%
    \restorebibmacro{cite}%
    \usebibmacro{cite}}}
\renewbibmacro*{textcite}{%
  \ifboolexpr{
    ( not test {\iffieldundef{prenote}} and
      test {\ifnumequal{\value{citecount}}{1}} )
    or
    ( not test {\iffieldundef{postnote}} and
      test {\ifnumequal{\value{citecount}}{\value{citetotal}}} )
  }
    {\DeclareFieldAlias{textcitehyperref}{noformat}}
    {}%
  \printtext[textcitehyperref]{%
    \restorebibmacro{textcite}%
    \usebibmacro{textcite}}}

\let\cite\parencite

\usepackage{bbding}
\usepackage[utf8]{inputenc} 
\usepackage[T1]{fontenc}    
\usepackage{hyperref}       
\usepackage{url}            
\usepackage{booktabs}       
\usepackage{nicefrac}       
\usepackage{microtype}      
\usepackage{amsmath}
\usepackage{graphicx}
\usepackage{tablefootnote}
\usepackage{multicol}
\usepackage[nameinlink]{cleveref}
\usepackage{bbm}
\usepackage{multirow}
\usepackage{soul}
\usepackage{float}
\usepackage{placeins}
\usepackage{wrapfig}
\usepackage{blindtext}
\usepackage{tablefootnote}
\usepackage{amsfonts}
\usepackage{colortbl}
\usepackage{mathtools,amssymb}
\usepackage{bm}
\usepackage{makecell}
\usepackage{caption}
\usepackage{capt-of}
\usepackage{array}
\usepackage{calc}      
\usepackage{caption}   
\usepackage{subcaption}  
\usepackage{xcolor,colortbl}
\usepackage[bottom]{footmisc}

\usepackage{xcolor}   
\usepackage{soul}     

\sethlcolor{yellow!20}

\definecolor{mydeepgreen}{RGB}{0, 100, 0}

\graphicspath{{figures/}}

\title{AgentX: Towards Agent-Driven Self-Iteration of Industrial Recommender Systems}

\renewcommand{\today}{}

\author{\large AgentX Team}

\begin{abstract}

Recommendation algorithm iteration is moving from an artisanal, engineer-bound process toward an industrialized research loop, but this transition remains blocked by a structural execution bottleneck: the idea-to-launch cycle still depends on human engineers to generate hypotheses, modify production code, launch A/B experiments, and attribute online results.
Innovation therefore scales linearly with headcount rather than compounding with evidence, compute, and accumulated experimental knowledge.
We present AgentX, a production-deployed multi-agent system that fundamentally restructures this production function.
AgentX operates as a self-evolving development engine: it autonomously generates, implements, evaluates, and learns from recommendation experiments at a scale and pace that no manual workflow can sustain.

The system orchestrates four tightly coupled stages in a closed loop.
A Brainstorm Agent synthesizes evidence from historical experiments, system architecture, data analysis, and external research into ranked, executable proposals.
A Developing Agent translates each proposal into production-ready code through repository-grounded generation and multi-dimensional reliability verification.
An Evaluation Agent conducts safe online rollout with guardrail-vetoed A/B judgment, converting both successes and failures into structured knowledge assets.
A Harness Evolution layer (SGPO) then distills execution trajectories into semantic-gradient updates that continuously sharpen the agents themselves---making the system not merely automated, but self-improving.

In a three-week Kuaishou App deployment across main-feed and life-service recommendation, three AgentX workers turned 374 ideas into 10 launchable rollouts---per-worker throughput doubled weekly via self-evolution, delivering 8$\times$ concurrency and 3.7$\times$ business value over a manual engineer, a 0.561\% user app-time gain, and over RMB 100M annualized revenue.
Beyond online strategy iteration, the same closed-loop principle extends to model-side research, enabling autonomous paper reproduction, module ablation, and cross-paper architectural composition.
These results establish that self-evolving, batch-scale agent-driven iteration is no longer hypothetical---it delivers compounding returns in production recommendation systems today.

AgentX points toward a new division of labor in recommendation research and development: one layer of engineers collaborates with the agent system to accelerate strategy and model iteration at scale, while another layer evolves the agent framework and its foundation models to continuously amplify the system's capability.
The trajectory data from every experiment fuels both---connecting daily execution to long-term intelligence growth.
For the first time, the linear addition of human effort can be rewritten into compoundable leverage.

\vspace{2\baselineskip}

\hfill\includegraphics[width=4cm]{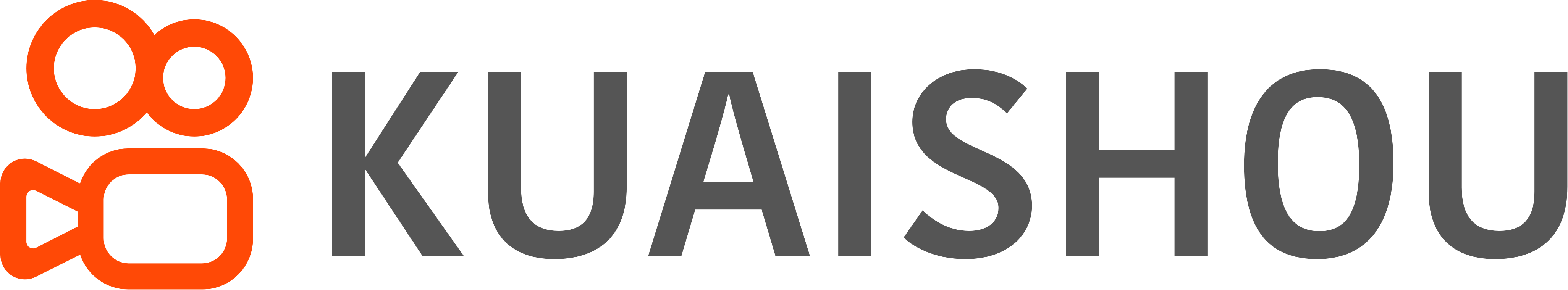}

\end{abstract}

\begin{document}

\maketitle

\newpage
\setcounter{tocdepth}{3} 

\tableofcontents

\newpage
\section{Introduction}

The iteration of industrial recommender systems~\cite{din, deepfm, onerec, rankmixer} has long relied on the continuous involvement of algorithm engineers. A complete algorithmic iteration typically traverses multiple stages, including data analysis, feature engineering, model development, online deployment, A/B testing, and attribution review, and further depends on heterogeneous data platforms, training frameworks, and monitoring tools, where the release cycle is often measured in weeks. Since industrial recommender systems are confronted with complex business objectives and stringent engineering constraints, effective system improvements heavily rely on expert experience that is difficult to encode explicitly. Consequently, the iteration throughput of the entire system is, to a large extent, constrained by the number of engineers and their cognitive load.

However, a substantial portion of engineers' effort is not spent on high-value judgments, but is dispersed across repetitive chores, such as data pulling, configuration modification, and pipeline maintenance. This dilutes the truly scarce capabilities, namely proposing high-quality hypotheses, understanding the causes of experimental failures, and identifying directions for system optimization, under substantial engineering friction. Conventional remedies, whether scaling up the engineer headcount, or building stronger underlying platforms, are essentially a linear amplification of the manual development process, and fail to alter the structural bottleneck in which recommender system iteration heavily depends on human labor and individual experience.

As large language models (LLMs)~\cite{qwen3, deepseekv3, gpt4, gemma2} continue to advance in code understanding, multi-step reasoning, error diagnosis, and tool invocation, driving algorithmic iteration with the agent as the executing entity has begun to be practically feasible. 
An agent~\cite{kimik2, glm5.0} is capable of not only comprehending the full engineering context such as feature definitions, training scripts, and experiment logs, but also acting directly upon it by modifying code and invoking internal platforms, thereby shifting the core bottleneck of recommender system iteration:
\begin{center}
\textit{From \textbf{the labor of algorithm engineers} to \textbf{the capability of agent systems}.} 
\end{center}
The former is heavily dependent on individual experience, whereas the latter is repeatable and compoundable. Once the execution trajectory of every online iteration can be recorded, analyzed, and fed back into the underlying agent framework,
the system itself is no longer merely an linear automation tool. 
Instead, it becomes a development mechanism that grows stronger over time, thereby making the construction of an automated development closed loop practically feasible.

Accordingly, in recent years, \textbf{LLM-driven automation of machine learning (ML) development} has witnessed a series of representative works, which can be characterized by three progressively higher levels of autonomy. 
The first level, \textbf{pipeline automation}, integrates LLMs into predefined AutoML~\cite{autoweka} and data science workflows, while leaving the task formulation and overall solution space largely specified by humans~\cite{caafe}. 
The second level, \textbf{experimental automation}, moves beyond executing established pipelines by allowing agents to explore implementations and experimental strategies, and to refine their decisions based on empirical feedback~\cite{automl-agent, mle-star}. 
The third level, \textbf{research automation}, further extends autonomy to the formulation and validation of research, where problems, hypotheses, methods, and evidence are jointly explored~\cite{ai-scientist, agent-laboratory}. 
These efforts collectively push machine learning development from single-stage assistance toward agent-based automation that integrates engineering execution, experimental iteration, and knowledge discovery.

Nevertheless, when we examine these works through the perspective of industrial recommendation scenarios, three common structural gaps emerge:
\begin{itemize}
    \item \textbf{Lack of real online feedback.} The success signals adopted by existing systems are predominantly drawn from offline metrics or human expert ratings. 
    While such signals are reasonable for academic evaluation, they are not equivalent to a positive contribution to real business value~\cite{ab-online}.  
    Online A/B results are the reward signal that is most aligned with real value and least susceptible to noise~\cite{ab-infra, ab-agent}, yet are largely absent from existing studies.
    \item \textbf{Limited validation at industrial scale.} The dataset, model, and pipeline scales validated in current public works still lag behind real-world recommendation systems, which involve billion-level samples, coupled multi-objective optimization, and multiple business lines running in parallel~\cite{moment-cross, onelive}. 
    Such scale discrepancies not only affect absolute performance, but further fundamentally reshape the bottleneck structure of the system, where engineering constraints that are negligible at small scale become critical variables that determine whether the closed loop holds at industrial scale.
    \item \textbf{Absence of continuous evolution mechanism.} 
    Most existing ML engineering agent systems do not accumulate the parameters, prompts, and workflows of the system after a single task is completed, resulting in the absence of long-term compounding effects.
    This results in the system struggling to adapt to new tasks and changing environments, repeatedly falling into the same pitfalls instead of reusing engineering experience, thereby restricting the continuous improvement of the system's intelligence level.
\end{itemize}

In other words, existing works have answered the question of \textbf{whether an agent can accomplish ML tasks in a controlled environment}, but have yet to address the more business-essential question of \textbf{whether an agent can continuously deliver online gains in real industrial recommender systems and self-evolve over time}.
Once agents render the generation and validation of ideas no longer scarce as a prior, the key shifts to how to construct a closed-loop and full-lifecycle development pipeline tailored to real business scenarios.

\begin{figure*}[t!]
\begin{center}
\includegraphics[width=\linewidth]{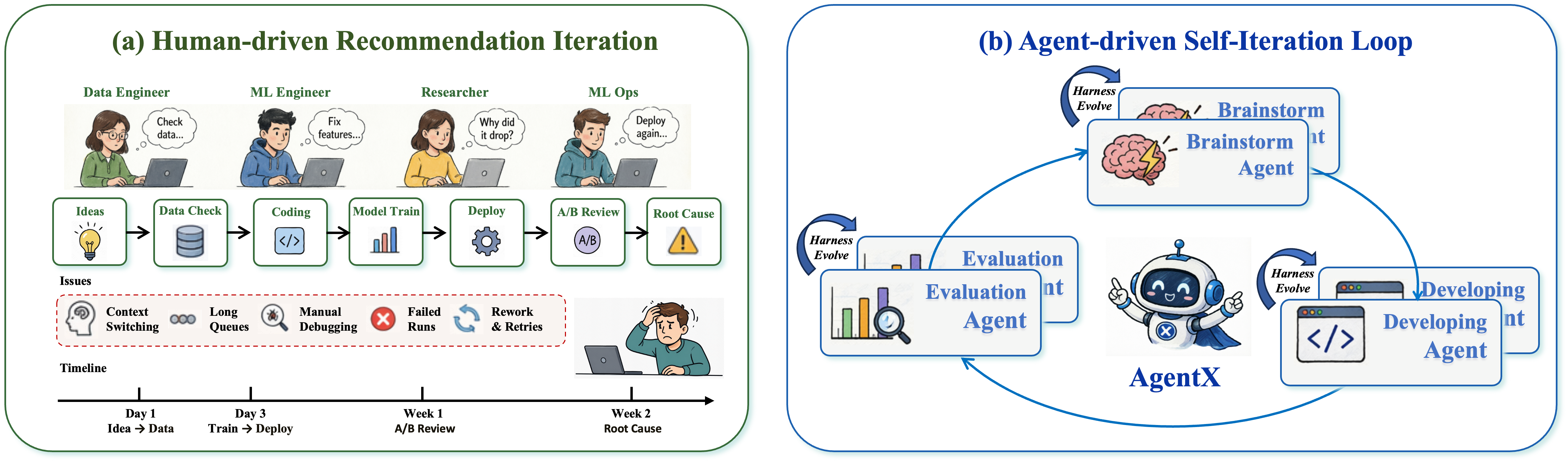} 
\caption{AgentX transforms recommendation iteration from a human-driven, manually handed-off pipeline into an agent-driven closed loop, where online A/B feedback and trajectory data continuously improve the system itself.}
\label{fig:intro}
\end{center}
\vspace{-0.3cm}
\end{figure*}

Motivated by the above background, we propose and deploy \textbf{AgentX}, an automated development framework tailored to industrial recommender systems. 
The objective of AgentX is not to let a LLM accomplish one-off ML tasks within a sandbox, but to continuously drive algorithmic iteration in real recommendation business. 
Specifically, AgentX forms a complete online feedback closed loop, encompassing idea generation, code and pipeline development, A/B analysis feedback, and the deduction of the next round. 
Every execution process is persistently distilled in the form of trajectories, which serve as the data source for subsequently optimizing agent workflows, training foundation models, and improving system policies. 
Therefore, AgentX is not merely a tool that automates manual steps, but a recommender development system endowed with self-iterative capability.

The main contributions of this work can be summarized as follows:
\begin{itemize}
    \item \textbf{Complete closed loop.} We have run through, in real industrial recommendation scenarios, an end-to-end pipeline of algorithmic development and evolution, spanning from investigation to production and deployment. The system does not rely on humans to stitch every step together. Instead, the agent continuously drives the process under unified orchestration.
    \item \textbf{Real industrial reward.} The ultimate success criterion of AgentX is grounded in real online A/B feedback, which leverages the cleanest training signals from users, is oriented toward the genuine demands of industrial scenarios, and directly aligns the optimization objective of the entire system with real business gains.
    \item \textbf{Scalable gains and a self-iterative flywheel.} AgentX is capable of generating multiple ideas concurrently and validating them through a unified quality funnel, thereby improving experiment throughput and the scale of online gains. Meanwhile, the trajectories produced by every experiment feed back into the agent orchestration framework and the foundation model, such that the system continues to grow stronger throughout long-term operation.
\end{itemize}

In the following sections, we first anchor the distinctions between AgentX and related works, and then elaborate on the design of this automated development closed-loop system. Subsequently, we demonstrate the trajectory-based self-iterative flywheel, followed by experimental results and real business gains at scale. Finally, we conclude this technical report by summarizing how AgentX reshapes the existing iteration paradigm of industrial recommender systems.

\section{Related Work}
\subsection{LLM-driven AutoML}

Traditional AutoML~\cite{autoweka} formulates machine learning development as optimization over human-defined model, pipeline, and hyperparameter spaces. LLMs extend this paradigm through semantic reasoning, executable code generation, and interaction with development environments. 
Recent work can be organized into three progressively higher levels of automation.

At the first level, \textbf{pipeline automation} incorporates LLMs into structured AutoML while still leaving the task definition, evaluation objective, and overall development procedure largely specified by humans. CAAFE~\cite{caafe} uses contextual knowledge to generate semantically meaningful features, whereas DS-Agent~\cite{ds-agent} and Data Interpreter~\cite{data-interpreter} combine planning, retrieval, and code execution to automate multi-stage data analysis. AutoML-Agent~\cite{automl-agent} further coordinate data processing, feature engineering, model construction, training, and verification within a unified workflow. 
Despite their broad coverage, these methods primarily aim to instantiate and execute a recognizable machine learning pipeline rather than autonomously determine a long-term experimentation strategy.

At the second level, \textbf{experimental automation} treats executable implementations and experimental trajectories as objects of autonomous search. ML engineering agents repeatedly modify code, execute experiments, interpret metrics and logs, and select subsequent actions according to empirical feedback. IMPROVE~\cite{improve} and MLE-STAR~\cite{mle-star} refine selected pipeline components based on attribution or ablation analysis, while AIDE~\cite{automl-agent} maintain multiple candidate solutions through structured search. R\&D-Agent~\cite{rd-agent} separates research reasoning from implementation, whereas AutoMind~\cite{automind} and MLEvolve~\cite{mlevolve} use memory or skill learning to accumulate experience across experiments.

At the third level, \textbf{research automation} further expands the decision space from solving a predefined task to formulating and validating the research itself. The AI Scientist~\cite{ai-scientist} and Agent Laboratory~\cite{agent-laboratory} integrate hypothesis generation, method development, experimentation, and scientific writing within a unified workflow, while more recent systems introduce structured exploration over alternative hypotheses, experimental designs, and supporting evidence~\cite{ai-scientistv2,ai-researcher}. Unlike ML engineering agents, whose primary objective is to improve a given metric, research agents must additionally reason about novelty, evidential validity, and reproducibility. Together, these three levels reflect a progression from automating established development procedures, to autonomously improving executable solutions, and ultimately to exploring machine learning research problems.

\subsection{Agentic Automation for RecSys}

Agentic automation for recommender systems concerns agents that design and improve the underlying recommendation models, rather than LLM agents that directly serve as recommenders. 
The early AutoML approaches~\cite{autofis, autoctr, nasrec} also focused on the spatial search of specific components within the predefined recommendation system framework.
Recent agentic approaches move beyond fixed operators by reasoning directly over executable implementations and experimental feedback.

A distinctive challenge in recommender development is that offline ranking metrics only imperfectly reflect online user responses.
Therefore, the industry's exploration in this direction will further connect model modification with production trials.
Self-EvolveRec~\cite{selfvolverec} evolves recommendation code using ranking metrics and diagnostic feedback to identify and address model failures.
Self-Evolving~\cite{self-evolving-recsys} Recommendation System integrates offline search with online validation, while AgenticRecTune~\cite{agenticrectune} optimizes configurations across multi-stage recommendation pipelines using reusable experience. 
A/B Agent~\cite{ab-agent} complements these approaches by simulating user interactions to provide richer evaluation signals beyond offline ranking metrics. 
However, reliably translating delayed, non-stationary, and potentially simulated user feedback into subsequent model improvements remains a central challenge.

\section{Multi-Agent Design Framework}

Before describing each agent and the harness evolution recipe in detail, we first outline the design principle of AgentX: what does a good iteration loop look like in an industrial recommender system? In classical LLM agent tasks involving math, code, or sandboxed ML benchmarks, it is usually assumed that the environment exposes a verifiable reward, that a single execution trace is sufficient to judge success, and that the optimization target is fixed before the agent starts. However, industrial recommendation departs from all three assumptions, since the reward signal lives in delayed online A/B feedback rather than offline metrics, a single deployment is gated by guardrails and human review before any score becomes available, and the optimization target itself drifts as business objectives, traffic mix, and platform constraints evolve. Therefore, a production-grade recommendation agent loop is closed-loop rather than one-shot. Specifically, it requires turning a vague business intent into evidence-grounded proposals, translating each proposal into repository-consistent code, validating the change through safe online rollout with guardrail veto, and feeding both positive and negative trajectories back into the system so that the loop itself improves over time. This distinction motivates our architectural and harness-evolution design.

We decompose AgentX into four stages along a single closed loop, where the first three jointly carry one idea from intent to validated online outcome, and the fourth feeds the resulting trajectories back to refine the loop itself:
\begin{itemize}
    \item \textbf{Brainstorm Agent.} Converting an under-specified user intent into a small, ranked set of executable experiment proposals via bounded exploration and evidence-weighted generation.
    \item \textbf{Developing Agent.} Translating each selected proposal into production-ready code through repository-grounded generation and a verification-oriented implementation loop.
    \item \textbf{Evaluation Agent.} Managing rollout and traffic assignment, judging A/B results with guardrail veto, and assetizing online outcomes into reusable reward signals and failure memory.
    \item \textbf{Harness Evolution.} Reasoning over accumulated execution trajectories to update individual subagent specifications via Semantic-Gradient-based Prompt Optimization (SGPO), admitted only through paired replay.

\end{itemize}

\begin{figure*}[t!]
\begin{center}
\includegraphics[width=\linewidth]{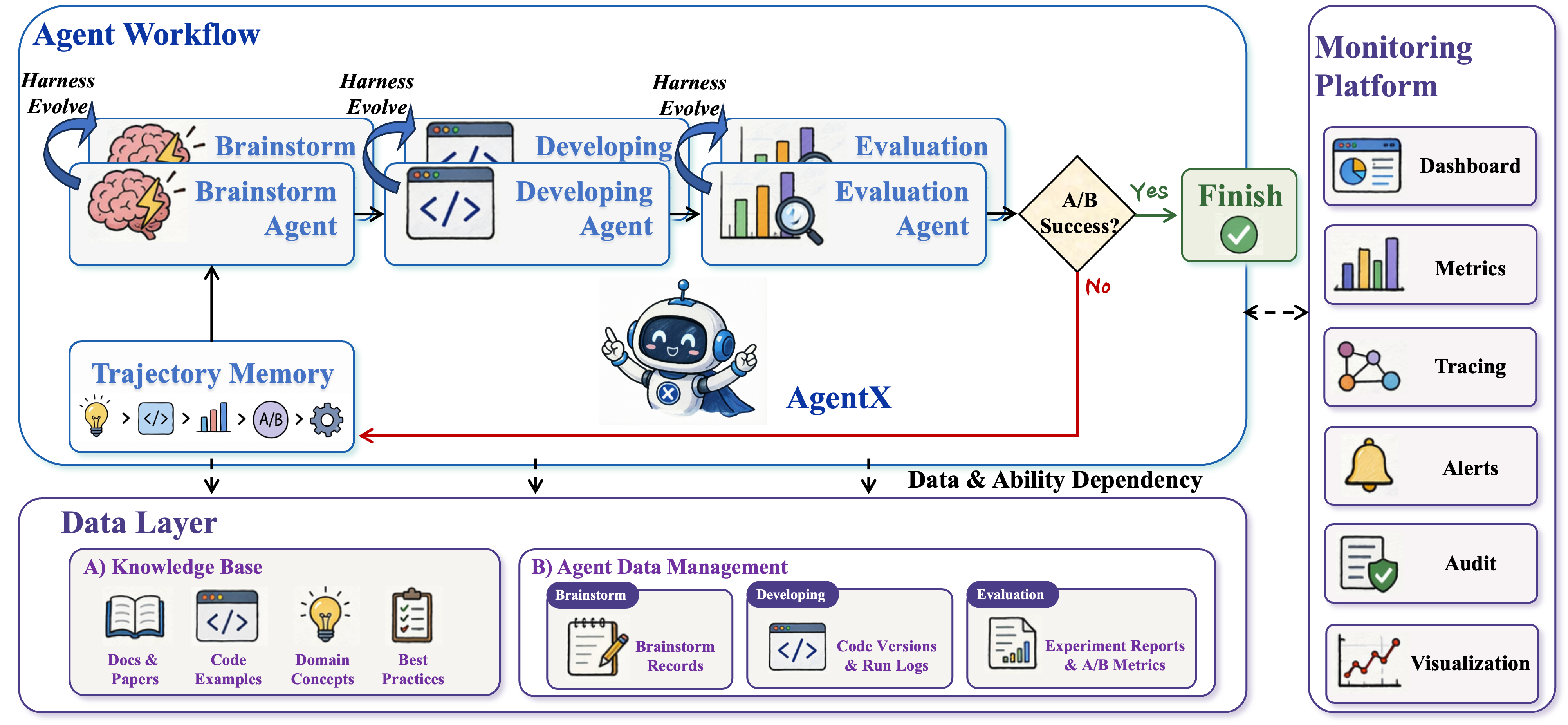} 
\caption{Overall framework of AgentX.}
\label{fig:overall_agentx}
\end{center}
\vspace{-0.6cm}
\end{figure*}

Around this loop, a shared Data Layer, encompassing a Knowledge Base of current experience in practice, together with Agent Data Management records of experiment reports, persists every artifact the loop produces, while the Monitoring Platform continuously observes system health through dashboards, metrics, tracing, alerts, audit, and visualization.
The overall framework of \textbf{Agentx} is illustrated in Figure \ref{fig:overall_agentx}.

\section{Brainstorm Agent}

The \textbf{Brainstorm Agent} serves as the entry point of the AgentX loop: it determines what downstream agents are allowed to implement, evaluate, and launch.
Its role is therefore not to generate a long list of plausible ideas, but to turn a vague optimization intent into a small, ranked set of executable experiment proposals.
Each proposal must be grounded in production evidence, scoped to an allowed change surface, and specified precisely enough to be coded, launched, and later diagnosed.

\begin{figure}[ht]
\centering
\includegraphics[width=0.7\textwidth]{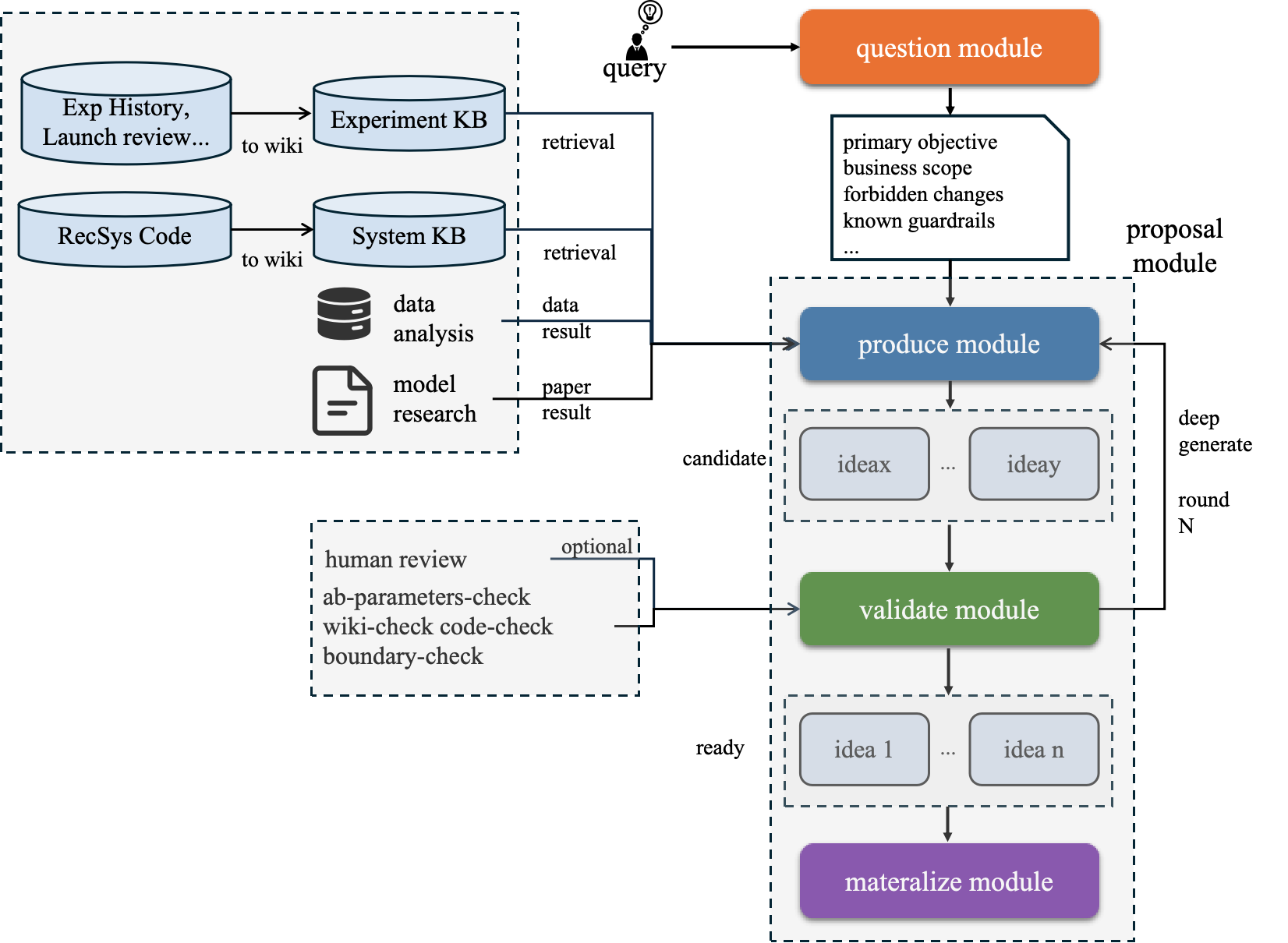}
\caption{Brainstorm Agent workflow. Historical experiment memory and system knowledge are organized into Experiment KB and System KB. The Brainstorm Agent clarifies the task, produces candidate ideas, validates them with review and boundary checks, and materializes accepted ideas for downstream execution.}
\label{fig:brainstorm_overview}
\end{figure}

This is difficult because industrial recommendation requests are rarely complete at input time.
A user may name only a loose objective; the relevant signals may carry fragile business semantics; historical launch reviews may already contain failed variants; and a proposal whose data source, pipeline stage, or metric path remains implicit can be silently mis-implemented or rejected at launch.
The Brainstorm Agent addresses this gap through the evidence-grounded proposal pipeline in Figure~\ref{fig:brainstorm_overview}.

This pipeline separates three responsibilities.
\begin{enumerate}
    \item \textbf{Question.}
    The question module turns an under-specified user intent into an explicit task boundary.
    \item \textbf{Idea Production.}
    The produce module generates candidate ideas under two coupled controls.
    \begin{enumerate}
        \item \textbf{Bounded proposal exploration.}
        Ideas are produced in batches rather than as a single free-form response, and each idea is tied to a target objective, possible implementation surface, expected downstream artifact, and maturity state.
        \item \textbf{Evidence context.}
        AgentX retrieves evidence from four sources: Experiment KB, System KB, Data Analysis, and Model Research.
        Experiment KB stores historical launch reviews and lessons; System KB stores architecture, feature, pipeline, and code-scope knowledge; Data Analysis provides metric and SQL-based empirical checks; Model Research grounds paper-derived mechanisms against production constraints.
    \end{enumerate}
    \item \textbf{Validation and Materialization.}
    The validate module filters candidates through human review and production checks, including A/B parameter checks, knowledge-base consistency checks, code-scope checks, and boundary checks.
    The materialize module then converts accepted candidates into structured proposal artifacts that can be handed off to validation, coding, launch, and later experiment diagnosis.
\end{enumerate}

\subsection{The Ambiguity Problem}
\label{sec:brainstorm:ambiguity}

The central difficulty of brainstorming is that the input is intentionally
under-specified while the output must be operationally precise.
If the agent explores too freely, it invents signals, assumes unavailable
features, or proposes changes outside the allowed code path.
If it is too conservative, it returns local variants of familiar strategies and
fails to expand the opportunity space.
The Brainstorm Agent is therefore designed as a boundary-setting and evidence
aggregation module: it preserves creative search, but only inside a recorded
production envelope.

For each task, the agent first normalizes the user request into an intake
boundary.
This boundary records the primary objective, allowed business scope, forbidden
changes, known guardrails, candidate output requirements, and unresolved
questions.
The purpose is not to force the user to specify everything upfront.
Instead, the boundary makes uncertainty explicit.
Fields that are known constrain the search; fields that remain unknown become
either conservative defaults or named follow-up probes.
This prevents ambiguity from being silently passed to the Developing Agent.

\subsection{Bounded Proposal Exploration}
\label{sec:brainstorm:bounded_exploration}

Once the boundary is fixed, the Brainstorm Agent explores proposals in batches
rather than through a single free-form response.
The batch structure is important because production ideas differ not only by
mechanism, but also by maturity.
Some ideas are ready to implement, some require a data or source probe, and
some are strategically interesting but depend on future platform or model
capability.
Treating these cases as the same type of output either discards useful
long-horizon directions or pushes immature ideas into coding too early.

AgentX therefore assigns each candidate to one of three states.
\textsc{Ready-to-implement} candidates have a concrete target, a named place
in the recommendation pipeline, a plausible objective path, and enough evidence
to enter review.
\textsc{Probe-first} candidates are promising but require a specific data,
source, or dry-run check before implementation.
\textsc{Moonshot-backlog} candidates preserve directions that may become useful
after future infrastructure, data, or model improvements.
This maturity split allows the system to explore broadly without confusing
exploration with commitment.

The batch loop is also residual.
After each round, rejected directions, historical duplicates, violated
constraints, and already-covered mechanisms are written into an avoid set.
Promising but incomplete directions are converted into explicit probes.
The next round then searches the remaining space instead of paraphrasing the
same ideas.
In this sense, the Brainstorm Agent is not merely sampling suggestions from an
LLM; it is contracting the feasible opportunity space under production
constraints.

\subsection{Evidence-Weighted Proposal Generation}
\label{sec:brainstorm:evidence_weighting}

Brainstorming is unreliable if all retrieved context is treated as equally
authoritative.
Different candidates need different kinds of evidence.
A novelty-oriented idea should be checked against historical launch reviews to
avoid repeating known failures.
A code-path-sensitive idea should be checked against architecture and
source-scope knowledge to avoid impossible handoffs.
A metric-diagnosis idea should be grounded in data analysis rather than
intuition.
The Brainstorm Agent therefore represents knowledge as a candidate-specific
evidence mixture, rather than as a fixed retrieval result.

Let $c$ denote a candidate idea and $q$ denote the structured intake boundary.
The evidence sources include four sources:
\begin{equation}
    \mathcal{K} = \{\text{Experiment KB}, \text{System KB}, \text{Data Analysis}, \text{Model Research}\}.
\end{equation}
Then a weighting mechanism lets the same candidate format support different reasoning regimes without hard-coding a single retrieval policy:
\begin{equation}
      \alpha_k(q,c) \geq 0,\qquad \sum_{k \in \mathcal{K}} \alpha_k(q,c)=1,
\end{equation}
where $\alpha_k(q,c)$ reflects how relevant and reliable source $k$ is for the current question and candidate.
Each source produces an evidence score $e_k(c)\in[0,1]$, and the weighted evidence term is
\begin{equation}
      E(c \mid q) = \sum_{k \in \mathcal{K}} \alpha_k(q,c)\, e_k(c).
\end{equation}
The final candidate score combines this evidence term with objective alignment,
business validity, implementation feasibility, handoff completeness, and risk:
\begin{equation}
  S(c \mid q)
  =
  \lambda_o O(c,q)
  + \lambda_b B(c)
  + \lambda_f F(c)
  + \lambda_h H(c)
  + \lambda_e E(c \mid q)
  - \lambda_r R(c).
  \label{eq:brainstorm_score}
\end{equation}
Here $O$ measures alignment with the user's primary objective, $B$ measures
business semantic validity, $F$ measures implementation feasibility, and $H$
measures whether the handoff is complete enough for downstream execution.
The risk term $R$ penalizes duplicate directions, unresolved core signals,
overly broad scope, unsafe trade-offs, and other proposal defects.

\subsubsection{Experiment KB}

\textbf{Experiment KB} stores historical launch reviews, business definitions, past experiment conclusions, and documented lessons from previous iterations. 
It captures not only whether an idea succeeded or failed, but also the context behind the result, such as the target scenario, affected user group, metric movement, launch decision, and post-hoc diagnosis.

This source is mainly used to prevent the Brainstorm Agent from rediscovering known failures or misusing business concepts. 
For example, a candidate that appears novel at the modeling level may have already been tested under a different name, or may violate a business definition that was clarified in a previous launch review. 
Experiment KB provides the historical grounding needed to detect such cases before downstream execution.

In the Brainstorm Agent, Experiment KB receives higher weight when a candidate depends on novelty, prior empirical evidence, or business semantics. 
It helps penalize duplicate directions, avoid previously falsified assumptions, and reuse lessons from successful launches. 
By incorporating historical experiment memory into candidate scoring, Experiment KB makes brainstorming cumulative rather than stateless.

\subsubsection{System KB}

\textbf{System KB} provides structured knowledge about the internal recommendation system, including model architecture, feature definitions, DSL behavior, pipeline boundaries, configuration semantics, and source-code scopes. 
Its purpose is to reduce hallucination when the Brainstorm Agent proposes code-path-sensitive ideas: a candidate should not only be semantically reasonable, but also implementable within the actual system.

System KB is built as a structured domain wiki rather than a flat document collection. 
It organizes private system knowledge into three layers: a schema layer that defines the fields and relations of each knowledge type, a wiki layer that stores structured Markdown entries following the schema, and a raw-source layer that links each entry back to its original code, configuration, or documentation for verification. 
This structure allows the agent to retrieve knowledge at the level of modules, features, pipelines, and implementation boundaries, instead of relying on coarse keyword search.

System KB is maintained through an ingest--query--lint lifecycle. 
During ingestion, multi-agent extractors convert heterogeneous sources, such as DSL definitions, C++ code, configuration files, and engineering documents, into standardized wiki entries. 
During query, schema-aware retrieval provides candidate-specific context to downstream agents, especially when an idea depends on where a change can be implemented or whether a required feature is available. 
During linting, incremental updates triggered by code diffs and periodic checkers keep the knowledge base consistent with the evolving system.

In the Brainstorm Agent, System KB is therefore weighted more heavily for feasibility-sensitive candidates. 
It helps reject ideas that violate pipeline boundaries, require unavailable features, depend on unsupported DSL behavior, or cannot be handed off to the correct code scope. 
By grounding brainstorming in verified system knowledge, System KB turns architectural and implementation constraints into explicit evidence for candidate scoring.

\subsubsection{Data Analysis}

\textbf{Data Analysis} provides empirical evidence for candidates whose validity depends on observed data patterns rather than intuition alone. 
It can access historical analysis reports, metric definitions, SQL plans, offline statistics, and current SQL query results when necessary. 
This makes it especially useful for ideas that rely on assumptions about user behavior, traffic distribution, feature coverage, sample bias, or metric decomposition.

For each candidate, Data Analysis helps verify whether the underlying empirical premise actually holds. 
For example, before proposing an idea that targets long-tail users, cold-start items, or a specific failure segment, the agent can check whether the segment is large enough, whether the metric drop is statistically meaningful, and whether the proposed direction matches the observed data pattern. 
This prevents the Brainstorm Agent from overfitting to anecdotal observations or proposing changes based only on surface-level metric movements.

In the Brainstorm Agent, Data Analysis receives higher weight when a candidate depends on metric diagnosis, segment-level behavior, distribution shift, or other data-grounded claims. 
It contributes evidence about whether the problem is real, whether the target segment is worth optimizing, and whether the proposed mechanism is consistent with observed data. 
By grounding candidates in empirical analysis, it turns brainstorming from intuition-driven proposal generation into data-aware hypothesis formation.

\subsubsection{Model Research}
\label{sec:research-agent}

\textbf{Model Research} provides external research evidence for candidates whose value depends on recent papers or prior academic findings.
Its role is to convert research papers into executable proposal knowledge, rather than treating them as plain text summaries.
This is important because a paper-level claim, such as a new gating mechanism or sequence module, may be valid in a controlled benchmark but infeasible or ineffective under the constraints of a production recommendation system.

To support this conversion, Model Research maintains a structured facts store for ingested papers.
Each paper is decomposed into typed claims, architecture components, and inter-paper relations.
Claims are annotated by their role, such as problem, assumption, method, finding, or limitation, and by evidence strength, such as isolated ablation, system-level evaluation, reported result, or author claim.
Architecture entries describe recommendation-specific components, including backbone, feature interaction operator, objective, and prediction head.
Inter-paper edges record whether papers extend, contradict, parallel, or apply each other.
This structure allows the agent to reason over papers at the level of reusable mechanisms instead of relying on coarse paper summaries.

The production baseline is represented in the same schema, together with its feature contract and training constraints, so that a paper-derived idea is grounded against the real system rather than evaluated in the abstract.
The feature contract enumerates every feature slot's name, dimension, and embedding-table identity; a proposal that preserves this contract can initialize directly from the production checkpoint and train only its newly added dense parameters.
The training constraints are hard: streaming online incremental training, no epochs, and a prohibition on techniques such as backbone freezing and early stopping.
This lets Model Research check whether a paper-derived idea preserves the required feature slots, fits the existing model architecture, and respects the production training rules; ideas that violate these constraints are rejected before being passed downstream.
The consequence is that the search space is constrained at the source by the actual production architecture and training regime, not merely evaluated against production metrics after the fact---the core industrial distinction from academic auto-research systems, where both the search space and the reward signal are synthetic.

In the Brainstorm Agent, Model Research is weighted more heavily when a candidate depends on external novelty or recent academic evidence.
It helps identify which paper claims are likely transferable, which modules are worth reproducing or ablating, and which combinations may be promising across papers.
By grounding paper knowledge against production constraints, Model Research turns recent research into feasible experimental proposals rather than unverified inspiration.
These proposals do not stop at single-paper reproduction: the same grounded knowledge drives a systematic exploration loop---reproduction, module ablation, and cross-paper composition---whose execution and self-evolution are described in Section~\ref{sec:developing-model-track} and Section~\ref{sec:evolve-model-track}.

\subsection{Validation and Implementation Handoff}
\label{sec:brainstorm:validation_handoff}

Candidate generation is followed by an admission step before any idea can
become a coding task.
The validator checks primary-objective alignment, business semantic correctness,
user constraints, model-score semantics, implementation feasibility, historical
overlap, A/B parameter feasibility, and maturity consistency.
This review produces both round-level and candidate-level decisions.
Only ready candidates that pass validation can enter the human approval gate;
probe-first ideas remain as explicit evidence-gathering tasks, and backlog
ideas remain outside the execution queue.

Human review is treated as a focused admission gate rather than as a repair
mechanism for every weak candidate.
When human review is required, the reviewer chooses among approve, revise,
defer, or reject over the machine-shortlisted directions.
For each approved idea, the implementation handoff creates exactly one formal experiment
record, a source manifest, and a handoff plan for the Developing.
The handoff contains the target behavior, required signals, expected metric
path, guardrails, and known implementation boundaries, but it deliberately does
not write production code inside the brainstorm stage.

This separation is what makes the module composable inside AgentX.
The Brainstorm Agent owns ambiguity reduction, evidence weighting, and proposal
admission.
The Developing Agent owns code realization and repository-level verification.
By forcing this boundary, AgentX prevents vague ideas from leaking into
implementation while still allowing the system to search beyond obvious local
changes.

\FloatBarrier

\section{Developing Agent}

The Developing Agent is the module that turns an approved proposal into a
\emph{verifiable code artifact}, along two parallel tracks that mirror how
recommender systems iterate in practice.

On the \textbf{online strategy} track, the artifact is a production code
change for feature- or strategy-level proposals, with the goal of safely
serving live traffic without silent reliability failures.
On the \textbf{offline model} track, the artifact is a training experiment
for model-architecture proposals, with the goal of producing conclusions
trustworthy enough to accumulate into a long-lived knowledge base of
model exploration findings.
A conclusion is deemed trustworthy only if four conditions hold simultaneously:
the implementation is verified to match the policy's declared causal mechanism
and expected observables; an isolated panel of expert agents reaches supermajority
agreement on the policy; metrics are extracted deterministically from raw
training logs rather than interpreted by an LLM; and any reported AUC gain
is backed by verified causal-chain attribution, such that unattributed gains
act as a brake signal rather than being recorded based on numerical improvement alone.

Despite their different deliverables, both tracks face the same fundamental
risk: promising ideas degrade into silent failures at the implementation stage.
Online, the target repository is large, internal conventions are partially
undocumented, and many errors evade compile-time checks: incorrect feature
names compile but read meaningless data, missing factory registrations disable
strategies without notification, and unguarded default-on changes can impact
live traffic prior to review.
Offline, the risk manifests analogously: hallucinated metrics and unverified
causal mechanisms can corrupt the knowledge base equally quietly, with no
explicit failure signal.

The Developing Agent is therefore designed not as a generic code generator, but
as a verification-oriented implementation system whose discipline applies
uniformly across both tracks.
The rest of this section details the online strategy track in
Section~\ref{sec:coding:online}, then adapts the same set of
principles to the offline model track in
Section~\ref{sec:developing-model-track}.

\subsection{Online Strategy Developing}
\label{sec:coding:online}
\subsubsection{The Production-Code Reliability Problem}

Coding in AgentX has a stricter contract than producing syntactically valid
patches.
The agent must preserve the proposal's intent, stay inside the allowed code
boundary, use only verified repository primitives, pass local and integration
checks, and leave a reviewable change that can be safely launched.
The core difficulty is that these requirements interact.
A patch can be logically aligned with the proposal but use a hallucinated
attribute; it can compile but miss a required pipeline registration; or it can
implement the strategy correctly but activate it without a default-off guard.

The main failure modes are therefore not generic programming errors.
They are repository-specific reliability failures.
Attribute hallucination arises when the agent invents fields in
the user feature schema, the context feature schema, or the item feature schema.
DSL misuse arises when it guesses ranking DSL operator names or argument contracts.
Harness-pattern violations arise when a change is placed in the wrong queue,
registered incompletely, or bypasses a required safety pattern.
Finally, every extra correctness-repair loop and every human intervention
reduces automation reliability, because the goal of AgentX is to close the
idea-to-launch loop with minimal manual rescue.

\subsubsection{Repository-Grounded Code Generation}
\label{sec:coding:grounding}

The Developing Agent handles these risks by grounding implementation in two
sources of repository knowledge.
The first is a project-specific knowledge base that records change patterns,
registration conventions, feature-switch rules, and examples of accepted
patches.
The second is a case toolbox: a set of deterministic tools and checkers that
force the agent to verify facts before using them.
Together, they convert the coding task from open-ended generation into a
sequence of grounded decisions.

\begin{figure*}[t!]
\centering
\includegraphics[width=0.82\linewidth]{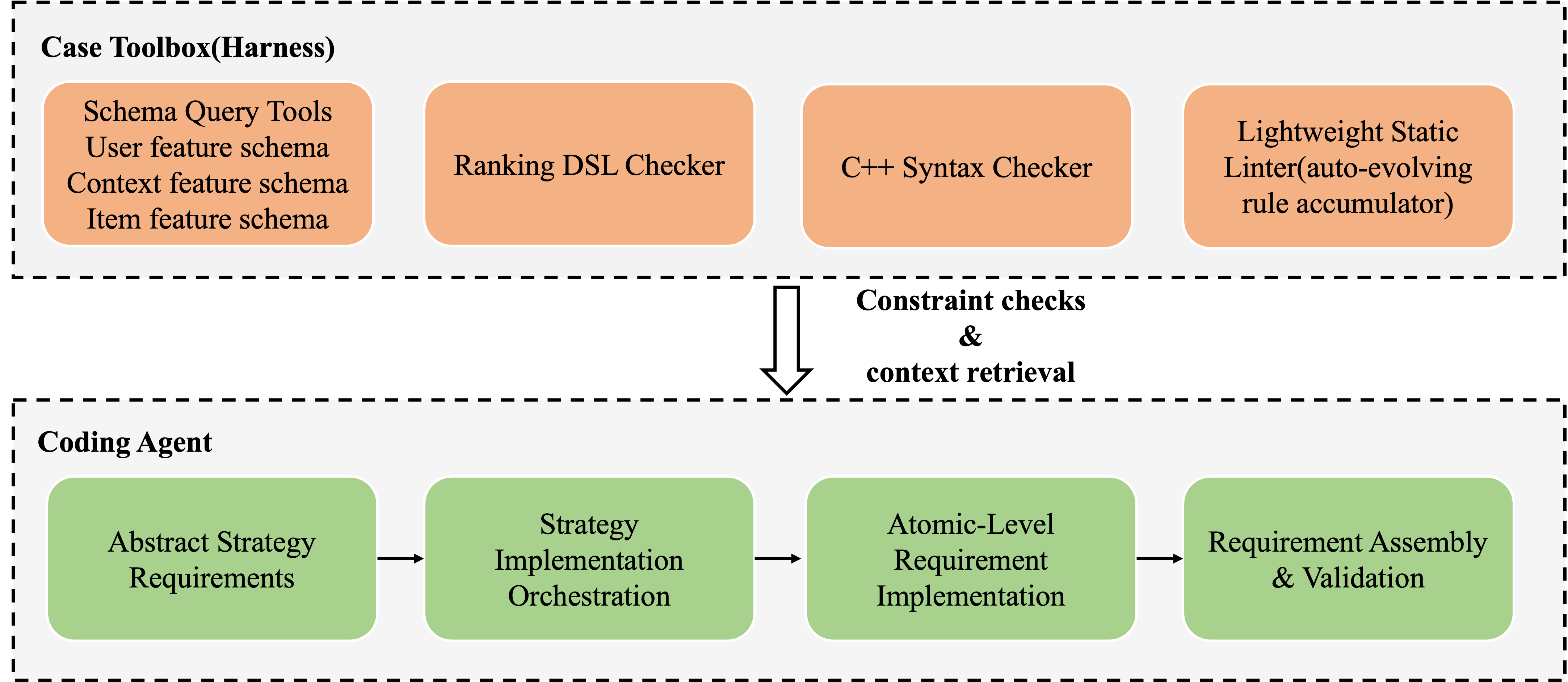}
\caption{Architecture of the Developing Agent for the online strategy track.
The central pipeline transforms a strategy specification into a validated code
submission through four sequential stages: abstracting requirements, orchestrating
the implementation plan, executing atomic sub-requirements, and assembling with
validation.
The case toolbox (right) injects constraint checks and context retrieval into the
pipeline, comprising schema-retrieval tools for user-, context-, and item-side
feature structs, a ranking DSL checker, a C++ syntax checker, and a lightweight
static linter that accumulates project-specific rules automatically.}
\label{fig:coding_overall}
\end{figure*}

The most important grounding rule is that feature attributes must be queried
before use.
Schema query tools for user-side, context-side, and item-side feature structs
return the available fields for each domain.
The agent is required to invoke the relevant tool before reading an attribute,
so field names become verified facts rather than language-model guesses.
For ranking DSL calls, a compiler-backed checker validates operator names,
argument types, and calling contracts.
For C++ idioms and project-specific syntactic sugar, a fast
lightweight static linter catches violations before the patch enters a
heavier build pipeline.

\subsubsection{Verification-Oriented Implementation Loop}
\label{sec:coding:verification_loop}

Each coding task follows a staged loop.
The agent first abstracts the approved proposal into an implementation plan:
which files may change, which signals are required, which pipeline stage owns
the logic, which feature switch guards the behavior, and which checks must
pass before submission.
It then implements atomic sub-requirements, assembles the patch, and runs
deterministic validation.
Failures are fed back as targeted repair instructions rather than as broad
regeneration prompts.

Two validation layers are especially important.
The \emph{accuracy loop} compares the implementation against the plan and
counts extra repair iterations as quality cost.
The ideal trajectory is that the first implementation already matches the
specified strategy.
The \emph{dryrun pipeline} then compiles and integration-checks the branch.
A clean trajectory passes Dryrun once; repeated Dryrun failures indicate that
the agent is relying on remote infrastructure as a debugging tool rather than
producing a locally disciplined patch.
Both layers are retained in the quality score below.

\subsubsection{Quality Scoring}
\label{sec:coding:scoring}

Coding quality is measured as a weighted reliability score rather than a
single pass/fail outcome.
Let $\mathcal{N}=\{1,\ldots,8\}$ denote the eight observed failure dimensions,
and let $s_i\in[0,1]$ be the normalized success score for dimension $i$.
The overall coding score is calculated as follows:
\begin{equation}
  Q_{\mathrm{code}}
  =
  \sum_{i\in\mathcal{N}} \lambda_i s_i,
  \qquad
  \sum_{i\in\mathcal{N}} \lambda_i = 1.
  \label{eq:coding_score_general}
\end{equation}
In the current implementation, the weights are instantiated as:
\begin{equation}
  Q_{\mathrm{code}}
  =
  0.06s_1 + 0.12s_2 + 0.22s_3 + 0.08s_4
  + 0.06s_5 + 0.18s_6 + 0.18s_7 + 0.10s_8 .
  \label{eq:coding_score}
\end{equation}

\begin{table}[ht]
  \centering
  \caption{Eight-dimensional coding quality score factors.}
  \label{tab:score_dims}
  \begin{tabular}{clcc}
    \toprule
    \textbf{Factor} & \textbf{Meaning} & \textbf{Severity} & \textbf{Weight} \\
    \midrule
    $N_1$ & C++ syntactic-sugar violations                 & B &  6\% \\
    $N_2$ & Harness pattern violations                     & A & 12\% \\
    $N_3$ & Attribute hallucinations                       & S & 22\% \\
    $N_4$ & Ranking DSL check corrections                  & A &  8\% \\
    $N_5$ & C++ syntax check corrections                   & B &  6\% \\
    $N_6$ & Correctness loop iterations                    & S & 18\% \\
    $N_7$ & Manual operator interventions                  & S & 18\% \\
    $N_8$ & Dryrun pipeline passes                         & A & 10\% \\
    \bottomrule
  \end{tabular}
\end{table}

The dimensions are normalized so that higher is better.
For count-based failures, $s_i$ decreases as the number of corrections or
violations increases.
Manual intervention is treated as a hard binary gate:
$s_7=1$ if no human intervention occurs and $s_7=0$ otherwise.
Dryrun is scored relative to the ideal of a single pass,
$s_8=\max(0,1-(N_8-1)/2)$.
The three severity-S dimensions, attribute hallucination, correctness-loop
overhead, and human intervention, together receive 58\% of the total weight
because they most directly threaten autonomous production reliability.

\subsubsection{Failure Modes}
\label{sec:coding:failure_modes}

The scoring dimensions correspond to concrete implementation failures.
Without schema grounding, the agent may read a plausible but nonexistent item
attribute; this can pass superficial code review while corrupting online
behavior.
For example, before schema tools were enforced, the agent could invent a field
such as \texttt{author\_live\_cpm\_boost} in an item attribute struct.
The code would look reasonable in context, but the correct field had to be
resolved through the item feature schema tool; otherwise the implementation would read
an invalid or unintended attribute.
Without DSL checking, it may invent an operator or pass arguments in an
unsupported form.
One recurrent pattern was the creation of plausible-sounding but unregistered ranking operators,
even though the registered operator contract required
a different scorer interface.
The DSL checker turns this from a late compilation or review failure into a
targeted local repair.
Without registration-pattern knowledge, it may implement a scorer but fail to
wire it into the serving pipeline.
In such cases the core scoring file may be correct, but the factory or queue
registration is missing, so the strategy silently does nothing online.
This is why file-coverage and harness-pattern checks are separate from ordinary
syntax checks.
Without feature-switch discipline, it may enable a strategy by default rather
than hiding it behind a disabled flag.
The expected pattern is to keep the base score unchanged unless a dedicated
feature switch is enabled; unconditional score multiplication is treated as a
safety-by-default violation.

The purpose of the Developing Agent is to make these failures observable before a
patch reaches production review.
Its contribution to AgentX is therefore not simply implementation throughput.
It provides the repository-level reliability layer that allows proposals from
the Brainstorm Agent and Model Agent to become safe, inspectable code changes.

\begin{figure*}[ht]
\centering
\includegraphics[width=0.82\linewidth]{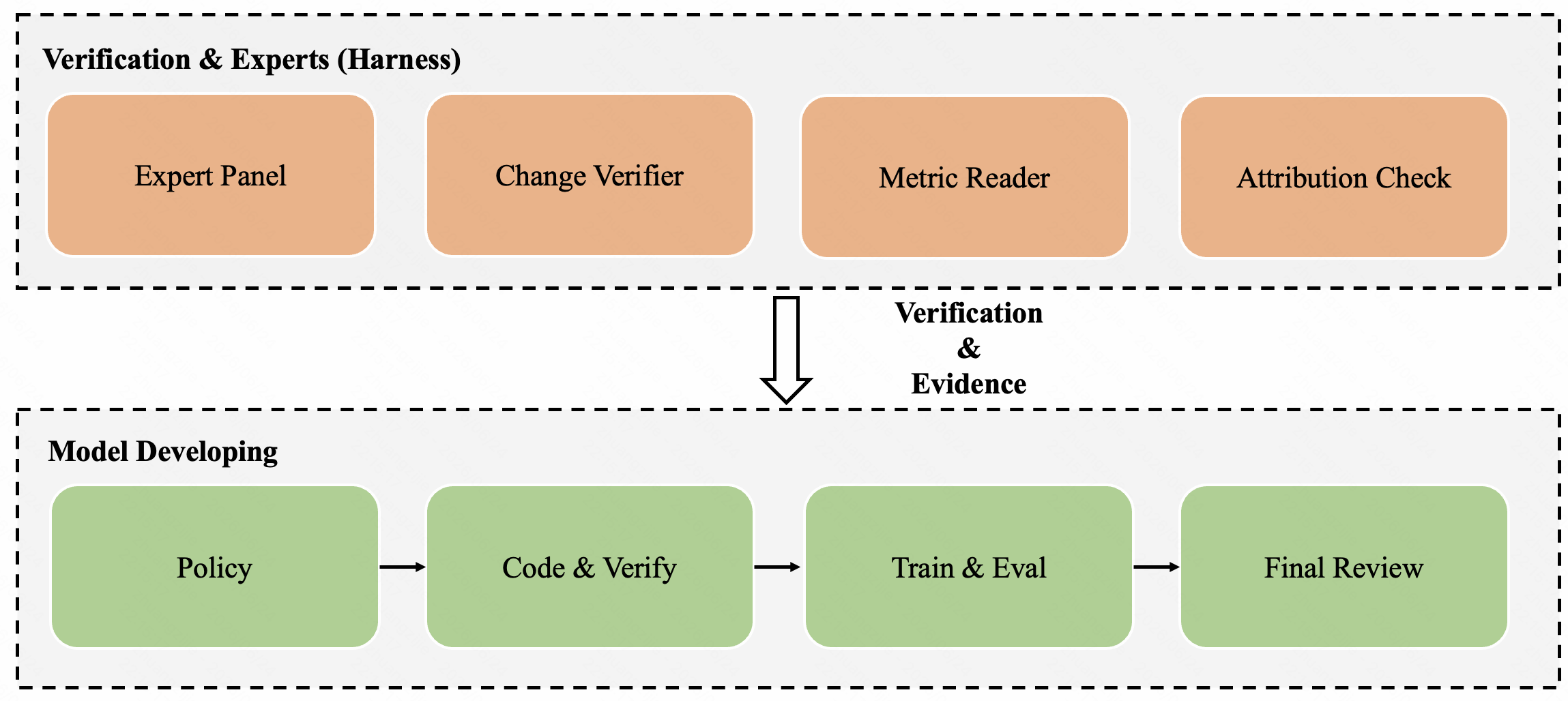}
\caption{Architecture of the Coding Agent for the offline model track. The central pipeline (bottom) turns a model-architecture proposal into a trustworthy training conclusion through four sequential stages: declaring a policy, implementing and verifying the code change, running training and evaluation, and reviewing the result with causal-chain attribution. The verification-and-experts harness (top) injects checks and evidence through an isolated expert panel voting for consensus, a change verifier that confirms the code matches the policy, a metric reader that extracts AUC from training logs, and an attribution check that adjudicates whether each causal-chain link holds.}
\label{fig:model_developing_overall}
\end{figure*}

\subsection{Model Developing}
\label{sec:developing-model-track}

A model-architecture proposal is developed differently from an online production change. The goal is not a safe, deployable patch but a trustworthy offline training experiment: the agent must implement the proposed architecture, run a real training-and-evaluation job on the production platform, and return a conclusion that can be believed. The chain of steps is long—interpret the proposal, edit model code, verify the change, submit a training job to a shared cluster, wait for results, parse training logs, and synthesize a verdict—and each step introduces a potential failure mode: LLMs can misinterpret directions, write subtly incorrect code, hallucinate metrics, or conflate correlation with mechanism; the training platform contributes GPU OOM errors, transient infrastructure faults, and log stalls. Over a batch of proposals running for days, any of these can corrupt the conclusions that feed back into the exploration memory. The execution agent is built around one answer to this problem: \emph{an LLM is permitted to be wrong only on judgment; every objective fact is produced by deterministic code.}

\subsubsection{Single-round pipeline}
Each round of the research loop follows the structure
\[
\texttt{policy} \;\to\; \underbrace{(\texttt{code} \;\leftrightarrow\; \texttt{verify})}_{\leq 3\text{ rewrites}} \;\|\; \underbrace{\texttt{experts} \times N}_{\text{concurrent}} \;\to\; \texttt{exec} \;\to\; \texttt{final\_review}.
\]
The round begins with \textbf{policy}, which declares not just what to change but also the claimed causal mechanism and a set of \emph{expected observables}—each named precisely so that the code agent can wire it into \texttt{tf.print} or \texttt{tf.summary}, with a description of what healthy vs.\ pathological behavior looks like (e.g., ``gate activation $>0.5$ after step~1000 indicates a live gate; $\approx 0$ indicates collapse''). This forces the round to commit upfront to how it will be falsified, rather than interpreting results post hoc. \textbf{Code} implements the policy within the designated file boundary. \textbf{Verify} then checks two things: that the resulting \texttt{git diff} semantically matches the policy direction, and that every declared observable name appears in the diff. If either check fails, the code agent rewrites—up to three attempts; exhausting the budget fails the round cleanly rather than proceeding with unverified code. Concurrently with this loop, \textbf{expert agents} evaluate the policy independently, each receiving only the policy text and its own private knowledge base, physically isolated from history and other experts' opinions. Expert consensus is computed by Python vote counting ($\geq \lceil 2N/3 \rceil$ for a supermajority), never by an LLM. \textbf{Exec} is a pure Python state machine—no LLM involved—transitioning through submission, polling, evaluation, and metric extraction, pulling AUC values from raw training logs via regular expression. The choice to exclude LLMs here is deliberate: metric extraction is a pattern-matching problem with a correct answer, and the cost of a hallucinated AUC number propagating into the knowledge base is high.

\subsubsection{Falsifiable attribution}
\textbf{Final review} synthesizes metrics, expert opinions, and training logs into a verdict. Critically, it also adjudicates each link of the declared causal chain individually—\texttt{verified}, \texttt{broken}, or \texttt{unclear}—by comparing training-log observations against the expected observables declared in the policy. Python then collapses this list deterministically: all links \texttt{verified} yields attribution status \texttt{CLEAR}; any \texttt{broken} or \texttt{unclear} link yields \texttt{UNCLEAR}. The verdict and attribution status are separate fields, and this separation matters: a round where AUC improves but attribution is unclear is a \emph{brake signal}—the system does not propagate an unattributed gain. An experiment on the RankMixer backbone illustrates why this constraint is necessary. Round~1 reproduced the paper's multiplicative gate $\mathbf{x} \cdot \tanh(\mathbf{V}_o \mathbf{x})$ with correct code and a small AUC gain of $+0.0003$. Without causal-chain enforcement, this would have been recorded as a success. But the declared observable for gate activation showed near-zero values throughout training—analysis confirmed that Glorot initialization of $\mathbf{V}_o$ produced $\mathbf{V}_o \mathbf{x} \approx \mathbf{0}$, causing $\tanh(\cdot) \approx 0$, zeroing the gate output and blocking gradients entirely. Attribution status: \texttt{UNCLEAR}. The gain was not recorded. Round~2 applied a one-line residual fix $\mathbf{x} \cdot (1 + \tanh(\mathbf{V}_o \mathbf{x}))$—which reduces to identity when $\mathbf{V}_o \approx \mathbf{0}$, restoring gradient flow—and returned $\Delta\text{AUC}{=}{+}0.0022$ with all causal-chain links \texttt{verified}.

\subsubsection{Robust execution under platform failures}
In a production training environment, many failures have nothing to do with the proposal itself: GPUs run out of memory, cluster jobs stall, infrastructure connections time out. If these are treated uniformly—either always retrying or never retrying—the system either wastes compute on fundamentally broken configurations or is killed by transient noise. The execution agent therefore classifies every failure before deciding how to respond. A pure-function classifier reads the log head (8~MB), log tail (256~KB), and up to 64 FATAL lines sampled from the middle, and maps the failure to a specific reason code. Deterministic errors—NaN gradients, feature-table conflicts, missing evaluation checkpoints—are abandoned immediately: retrying them cannot help. Transient faults—log stalls detected by fingerprinting the tail byte count and hash, infrastructure aborts—retry once. LLM gateway failures rotate to the next available gateway. The classifier evaluates reason codes in fixed priority order; an infrastructure symptom such as \texttt{ps\_aborted}, which can mask an underlying deterministic error, is always evaluated last, preventing a schema bug from being misclassified as a recoverable transient. A detached watchdog daemon applies this same policy to all active runs in a batch, self-healing stuck or failed runs without human intervention. Every abandoned run records its reason code explicitly (e.g., \texttt{given\_up:nan\_train}), so root cause is immediately readable. This execution-layer robustness is invisible during normal operation but essential for a system running dozens of parallel experiments over multiple days.

\section{Evaluation Agent}
The \textbf{Evaluation Agent} closes the production loop of AgentX: it determines whether a code change materialized by the Developing Agent should be kept, rolled back, or fed back as a negative lesson for the next round of iteration. Its role is therefore not to merely report A/B numbers, but to convert noisy, delayed, and partially observable online traffic into a trustworthy reward signal that the rest of the system can act upon. To this end, the Evaluation Agent is designed as an end-to-end pipeline rather than a single judgement module, so that real user feedback is consistently turned into the authoritative reward signal of AgentX and into reusable constraints that shape its future iterations.

\begin{figure*}[ht]
    \centering
    \includegraphics[width=0.8\linewidth]{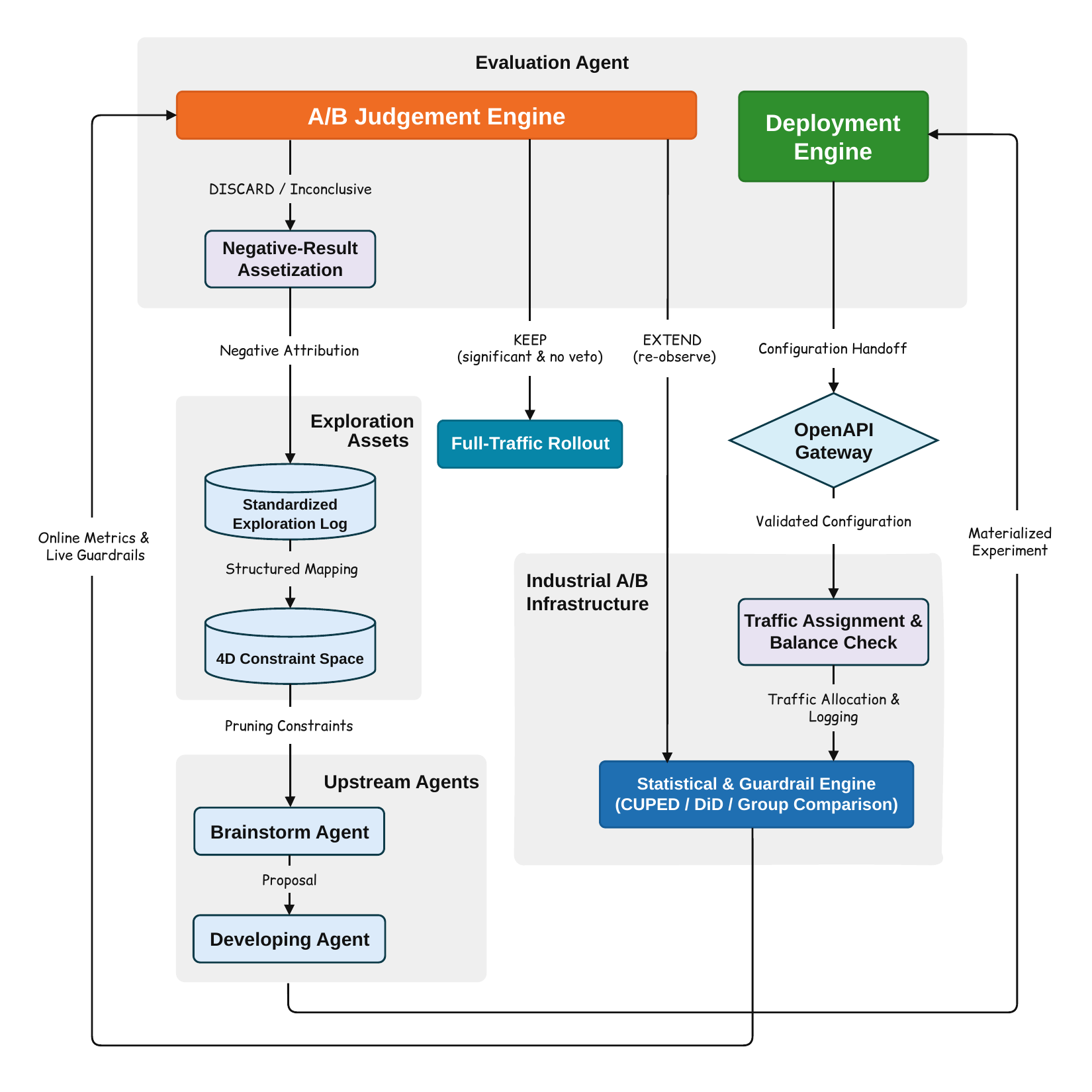}
    \caption{Architecture of the Evaluation Agent, which chains
    OpenAPI-mediated safe deployment, online A/B execution, and guardrail-vetoed
    judgement, turning production feedback into the reward signal for later AgentX
    iterations.}
    \label{fig:analysis_agent_arch}
\end{figure*}

\subsection{The Real-World Reward Problem}
\label{sec:analysis:reward_problem}

After the Brainstorm Agent proposes an experiment and the Developing Agent
materializes it, the system still does not know whether the change is worth
keeping.
Offline proxies and self-reflection are insufficient in recommender systems:
a strategy may improve an internal score while hurting long-term user
experience, or appear promising in a small slice while damaging a guardrail
metric.
AgentX therefore treats online A/B feedback as the authoritative reward signal
for system iteration.

This creates a second reliability problem.
The agent must launch experiments safely, allocate traffic without contaminating
other experiments, read noisy online metrics, apply strict rollout criteria,
and convert negative results into reusable memory.
The Evaluation Agent is designed around this full chain.
It is not just a reporting module; it is the mechanism that turns real user
feedback into decisions and future constraints.

\subsection{Safe Deployment and Traffic Assignment}
\label{sec:analysis:safe_deployment}

Before an experiment can produce reward, it must enter traffic safely.
Industrial recommendation systems usually route requests through multiple
business domains, layers, worlds, bucket ranges, and split factors.
Launching an automatically generated strategy without respecting this topology
can cause parameter conflicts, cross-experiment contamination, or inconsistent
user assignment.
The Evaluation Agent therefore begins with deployment addressing.

The deployment layer maps each experiment to the correct business domain and
world, chooses an appropriate split factor, and allocates mutually exclusive
traffic buckets.
For account-bound experiments, routing can be based on user ID; for
device-side experience changes, it can use device ID; for mixed populations,
an UID-first policy preserves consistency for logged-in users while falling
back to device identity for anonymous traffic.
When bucket assignment is underdetermined, the agent performs pre-experiment
balance checks and selects traffic groups that minimize baseline mismatch.

Safety controls are applied before rollout.
Parameter changes must pass an engineering whitelist so that an agent cannot
modify unauthorized scheduler or serving controls.
Configuration rollout follows a canary path: after the API submits a
change, the system observes a minimum gray-release window before escalating to
full traffic.
If monitoring detects instability, the rollout is halted before the experiment
becomes a large-scale online risk.

\subsection{A/B Judgement with Guardrail Veto}
\label{sec:analysis:ab_judgement}

Once traffic is live, the agent must decide whether an observed effect is
trustworthy.
Online data is noisy, delayed, and sometimes partially missing.
The Evaluation Agent therefore separates metric extraction from
decision logic.
When pre-experiment history is available, it uses variance-reduction methods
such as CUPED.
When the data environment does not satisfy those assumptions, it falls back to
more robust comparisons such as difference-in-differences or direct group
comparison.
If upstream data is missing, the agent shrinks or shifts the observation window
instead of treating an incomplete query as a conclusion.

The decision policy is intentionally conservative, but its guardrail design
follows three principles that keep false negatives under control.

First, \textbf{guardrails are business-scoped, not universal}.
Each business domain (e.g., consumption, live streaming, e-commerce,
advertising) operates its own set of core metrics and veto thresholds,
calibrated to its specific optimization objectives.
An experiment in one domain is not subject to the full union of all guardrails
across the system; it is primarily accountable to the guardrail indicators of
the domain it operates in and to a small set of cross-domain stability metrics.

Second, \textbf{guardrails use composite economic-exchange metrics rather than
single-indicator vetoes}.
Instead of blocking on any individual metric crossing a hard line, the system
computes an aggregated lifetime-value (LT) exchange score that weights several
business objectives into a unified summary.
A single guardrail indicator moving negatively does not automatically trigger
a veto; the agent looks at the composite picture, which substantially reduces
noise-driven false negatives while still catching cases where a strategy
genuinely harms the ecosystem.

Third, \textbf{thresholds are designed as attention signals, not absolute
blocks}.
The primary role of guardrail thresholds is to flag risks for human review,
not to mechanically discard experiments.
When a guardrail is triggered, the strategy is escalated rather than
automatically discarded.
In practice, a high-benefit strategy that triggers a moderate guardrail
deterioration can still proceed through an exception-review path, provided
the gain is large enough to justify the externality.
This tiered design---hard block for severe deterioration, escalation for
moderate signals, and monitoring-only for observation metrics---prevents the
system from being excessively conservative while maintaining safety.

Together, these principles ensure that the guardrail veto catches genuinely
harmful strategies without systematically suppressing beneficial experiments.
A candidate is eligible for KEEP only when the primary objective clears both a
minimum effect threshold and statistical significance, and when guardrail
assessments confirm no unacceptable cross-domain damage.

The output of the analysis stage is therefore not a single number.
It is a structured verdict: KEEP, EXTEND, or DISCARD, together with the primary
effect, guardrail status, statistical method, observation window, and caveats.
This structured verdict is the reward record consumed by later stages of the
AgentX loop.

\subsection{Negative-Result Assetization}
\label{sec:analysis:negative_assetization}

Most production experiments do not become successful rollouts.
For AgentX, this is not wasted work.
A negative result is useful if the system records why the strategy failed and
prevents future agents from repeating the same path.
The Evaluation Agent therefore converts DISCARD and inconclusive
outcomes into exploration assets.

Each failed experiment is written with its root cause: missing significance,
guardrail deterioration, traffic mismatch, implementation-side caveat, or
business-context mismatch.
The record is also indexed by pipeline stage, business objective, affected
user or content segment, and strategy lever.
Before the next brainstorm or analysis cycle, AgentX can retrieve these assets
to avoid rediscovering failed directions.
For example, if a previous boost strategy for low-frequency users damaged a
retention guardrail in rough ranking, the same coordinate combination becomes
a high-risk region for future proposal generation.

This assetization step completes the closed loop.
Online A/B results decide which changes survive, but negative results decide
which future branches should be pruned or probed more carefully.
The Evaluation Agent therefore supplies both the reward signal and the
failure memory that make AgentX's self-iteration grounded in real production
feedback rather than model self-assessment.

\section{AgentX Harness Evolution}
Production experiment analysis captures only one side of the AgentX feedback loop. Although the Experiment Analysis Agent successfully diagnoses online outcomes, records failed directions, and translates A/B feedback into reusable memory, it leaves a critical gap. It does not explain why an upstream agent failed to capture the right constraints, missed a business-causal chain, produced an incomplete handoff, or generated code that violated repository conventions. To bridge this gap, AgentX introduces a second optimization layer over execution trajectories. The target is not the recommendation strategy itself, but the harness controlling how each subagent reasons, asks questions, validates evidence, hands off work, and recovers from failure.

We use the term \emph{harness} in a strictly constrained sense. While the full operating harness includes top-level orchestration, memory, tool interfaces, inter-agent protocols, and subagent instructions, the current production-safe version updates only one subagent-level harness specification at a time. During an evolution run, the foundation model, tool interface, top-level orchestration, and other subagents remain fixed. Only the target subagent's instructions, validation rules, output contract, and tool-use discipline are edited. This design ensures updates are fully inspectable and makes old-vs-new replays meaningful: any score change can be isolated to a local harness edit rather than a sweeping system rewrite. For this constrained setting, we introduce \textbf{Semantic-Gradient-based Prompt Optimization} (SGPO), an offline harness-evolution method that turns accumulated execution traces into local subagent prompt updates and admits them solely through paired replay. In short, SGPO treats natural-language diagnoses of trace failures as semantic gradients that revise a single subagent specification while the rest of AgentX remains fixed.

\subsection{Semantic-Gradient-based Prompt Optimization}

\subsubsection{Semantic-Gradient-based Prompt Optimization I}
\label{sec:harness:sgpo_i}

SGPO-I uses session traces as the evidence source.
Let $h_{t,i}$ denote the current harness specification of target subagent $i$ at evolution round $t$.
For each round, SGPO samples a small batch of traces related to subagent $i$ from the accumulated online trace pool.
The evaluator does not read the full session verbatim.
Instead, it extracts compact rubrics from the initial user query and later user inputs, because these fields usually contain the explicit task constraints and the implicit constraints revealed during interaction.

The first step is loss calculation, which produces the semantic gradient used for harness refinement.
Given the current harness, sampled trace evidence($\mathcal{T}$), and extracted rubrics($\mathcal{R}$), an evaluator agent($E_{\text{agent}}$) first writes a natural-language loss report and then condenses it into the semantic gradient $g_{t,i}$:
\begin{equation}
\ell_{t,i},\; g_{t,i}
=
E_{\mathrm{agent}}\!\left(h_{t,i};\ \mathcal{T},\ \mathcal{R}\right).
\label{eq:sgpo_update}
\end{equation}
The semantic gradient $g_{t,i}$ is not a numerical derivative.
It is a structured diagnosis of missing constraints, weak step ordering, underspecified evidence requirements, or incomplete downstream contracts.

The second step is semantic gradient update.
A refinement agent($R_{\text{agent}}$) converts this gradient into a local harness edit and produces the candidate revised harness:
\begin{equation}
h'_{t,i}=R_{\mathrm{agent}}\!\left(h_{t,i}, g_{t,i}\right).
\label{eq:sgpo_gradient_update}
\end{equation}
The refinement is limited to the target subagent's instruction, validation rule, output contract, or tool-use discipline; it does not rewrite the full AgentX harness.

After $h'_{t,i}$ is generated, SGPO immediately evaluates the old and new harnesses on the same replay tasks.
The replay task set is generated from the same trace pool: an LLM rewrites user queries and later user inputs into standalone user tasks.
Each replay task preserves the business domain, objective, guardrails, allowed change type, expected artifact, and known constraints while removing dependence on earlier dialogue context.
AgentX is then rerun with the old harness $h_{t,i}$ and the candidate harness $h'_{t,i}$ under identical tasks, and an evaluator scores both outputs using the same rubric family:
\begin{equation}
\begin{split}
\Delta J_i 
&= \operatorname{ReplayScore}(h'_{t,i})
 - \operatorname{ReplayScore}(h_{t,i}),\\
h_{t+1,i} 
&\mathrel{:=} h'_{t,i}, 
\quad \text{if } \Delta J_i > \epsilon 
\land \operatorname{Safe}(\Delta h_i).
\end{split}
\label{eq:sgpo_accept}
\end{equation}
Here $\Delta J_i$ is the paired-replay improvement and $\epsilon$ is the admission threshold.
If $\Delta J_i>\epsilon$ and the safety check preserves schema compatibility, tool boundaries, and human-review constraints, the candidate update is accepted and the updated state $h'_{t,i}$ serves as $h_{t+1,i}$ in the subsequent iteration.
Otherwise, the target subagent is not updated, and the rejected patch, evaluator score, and failure explanation are retained as refine experience for later rounds.

\begin{figure}[H]
\centering
\includegraphics[width=0.80\textwidth]{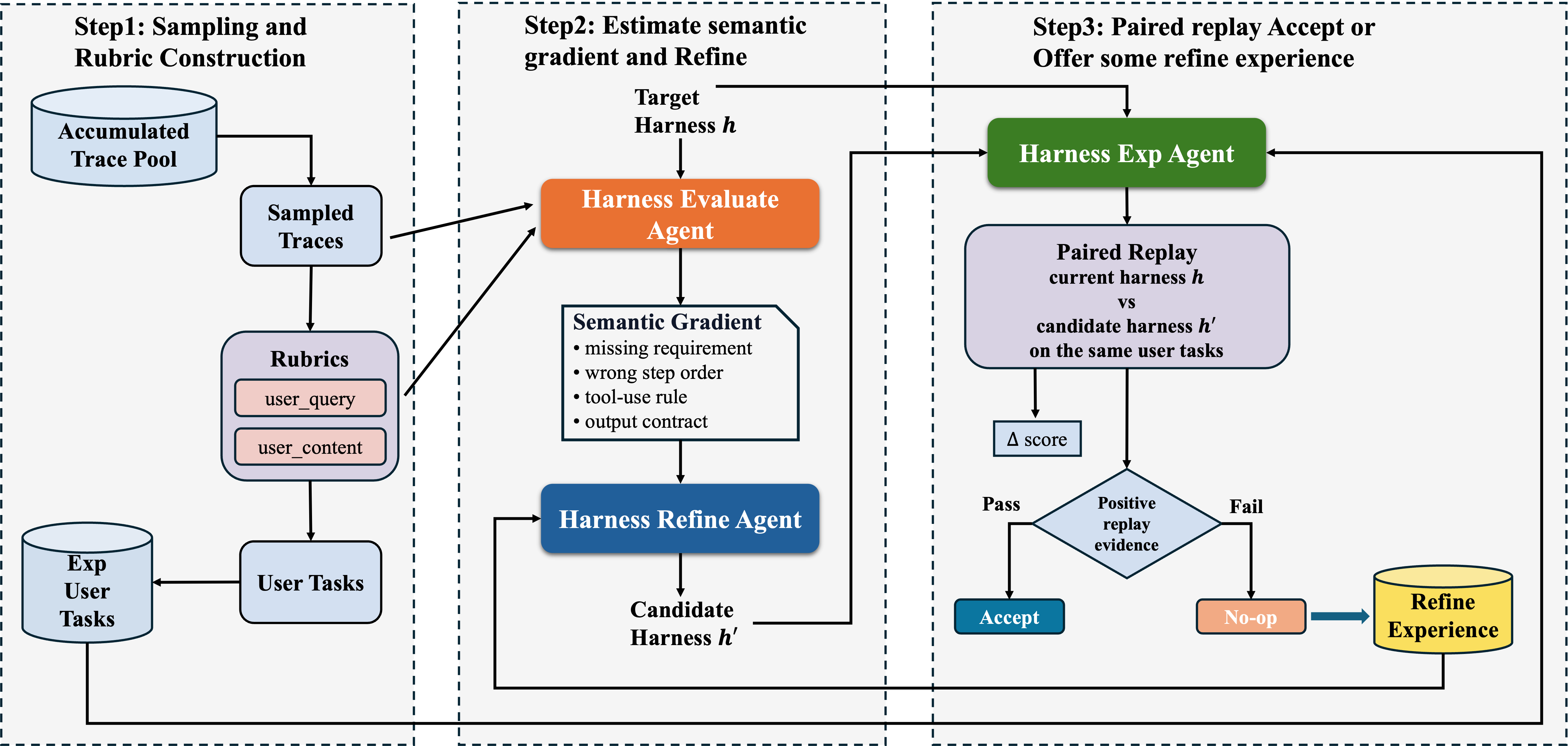}
\caption{SGPO-I trace-based harness evolution. Accumulated traces are sampled into rubrics and replay tasks; evaluator feedback is converted into a semantic gradient, refined into a candidate harness, and admitted only through paired replay.}
\label{fig:harness_offline_evolve}
\end{figure}

This optimization pipeline maintains clear inspectability and strict production safety through a three-stage closed loop (as illustrated in Figure~\ref{fig:harness_offline_evolve}).

\begin{enumerate}
    \item \textbf{Sampling and Rubric Construction (Step 1):} The cycle begins by drawing execution traces from the \textit{Accumulated Trace Pool} to form \textit{Sampled Traces}. Concurrently, historical evaluation tasks from the \textit{Exp User Tasks} database are used to construct explicit evaluation rubrics (e.g., \texttt{user\_query} and \texttt{user\_content}).
    
    \item \textbf{Semantic Gradient Estimation and Refinement (Step 2):} To calculate the linguistic ``loss,'' the \textit{Harness Evaluate Agent} analyzes the failures within the sampled traces against the current \textit{Target Harness} ($h$). This diagnosis yields a structured \textbf{Semantic Gradient}, which explicitly maps execution flaws such as missing requirements, incorrect step orders, tool-use violations, or broken output contracts. Acting as a prompt-space optimizer, the \textit{Harness Refine Agent} leverages this gradient to propose an updated \textbf{Candidate Harness} ($h'$).
    
    \item \textbf{Paired Replay and Experience Assimilation (Step 3):} To strictly gate production deployment, the \textit{Harness Exp Agent} conducts a \textbf{Paired Replay} evaluation; specifically, it executes both the current harness $h_{t,i}$ and the candidate harness $h'_{t,i}$ over identical user tasks. The \textit{Harness Exp Agent} evaluates the performance delta ($\Delta \text{score}$). If the replay yields positive evidence ($\Delta \text{score} > 0$), the candidate harness is formally \textbf{Accepted} into production. Conversely, if the update fails to improve performance, a \textbf{No-op} is triggered, and the failure mode is archived into the \textit{Refine Experience} database to guide subsequent optimization iterations.
\end{enumerate}

In our one production use case, SGPO-I is applied to the idea-generation stage, where failures often appear as vague objectives, missing evidence grounding, duplicate directions, or artifacts that downstream validation and coding agents cannot directly consume.
Table~\ref{tab:sgpo_brainstorm_rounds} summarizes one five-round evolution run over a brainstorm subagent.
The normalized replay score improves from $75.15\%$ to $98.00\%$.
The most important accepted edits are not generic prompt polishing; they are concrete contract changes, such as making the task contract explicit before idea generation and requiring each candidate to expose a business-causal chain.

\begin{table}[ht]
\centering
\caption{SGPO-I evolution trace for the brainstorm subagent. Each round samples traces, proposes a local harness edit, and admits the edit only after paired replay.}
\label{tab:sgpo_brainstorm_rounds}
\resizebox{0.85\textwidth}{!}{%
\begin{tabular}{clp{0.47\textwidth}cc}
\toprule
\textbf{Round} & \textbf{Edit focus} & \textbf{Harness implication} & \textbf{Avg. score} & \textbf{Normalized} \\
\midrule
1 & Task contract normalization & Restate business domain, target metric, scope, human-review requirement, hard constraints, and missing information before generating candidates. & 3.76 & 75.15\% \\
2 & Evidence grounding & Require an explicit evidence index and mark unavailable evidence instead of silently relying on intuition. & 3.89 & 77.82\% \\
3 & Candidate-quality schema & Standardize each idea around problem source, business-causal chain, strategy formula, implementation point, dry-run checks, and risk. & 4.53 & 90.67\% \\
4 & Novelty and rejection & Separate same-theme-but-new-point ideas from duplicated optimization points, and record rejected directions with reasons. & 4.87 & 97.33\% \\
5 & Evaluable handoff & Produce a candidate overview, priority order, validation/coding fields, unsupported items, and evaluation criteria for each idea. & 4.90 & 98.00\% \\
\bottomrule
\end{tabular}%
}
\end{table}

\begin{figure}[H]
\centering
\includegraphics[width=0.9\textwidth]{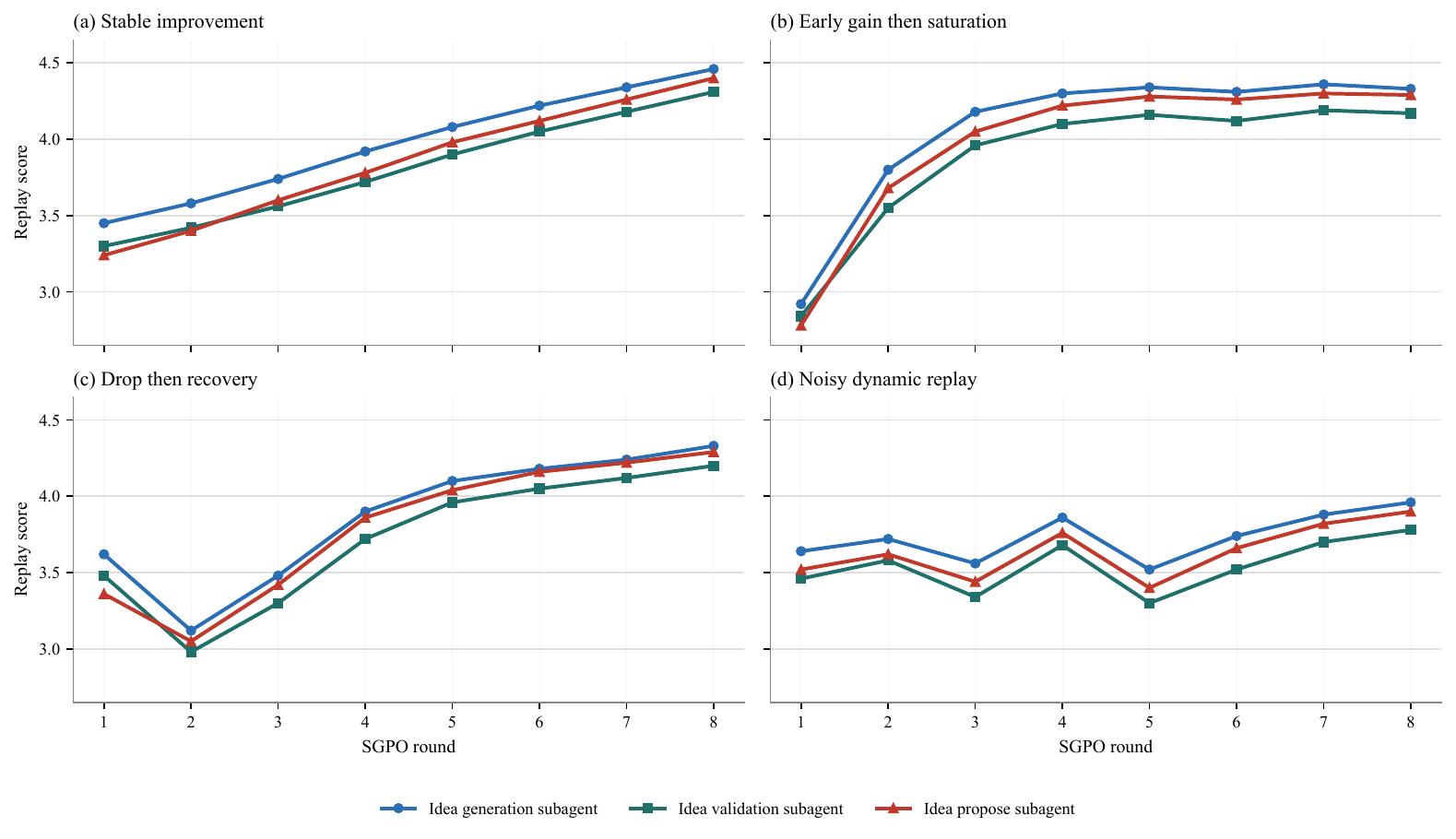}
\caption{Representative SGPO-I evolution patterns on the brainstorm workflow. Each panel shows eight SGPO rounds under a fixed setting: trace batch size 5, 8 rubric items, and 10 replay tasks. The three curves track the idea generation subagent, idea validation subagent, and idea propose subagent. Since rubrics and replay tasks are regenerated dynamically, harness evolution may show steady gains, early saturation, temporary regression, or noisy recovery rather than a strictly monotonic trend.}
\label{fig:sgpo_parameter_sensitivity}
\end{figure}

We further examine SGPO-I within the brainstorm workflow under a fixed operating setting.
Figure~\ref{fig:sgpo_parameter_sensitivity} summarizes four representative evolution patterns observed in the harness-evolution history.
Each panel contains eight SGPO rounds and tracks the idea generation subagent, idea validation subagent, and idea propose subagent.
The curves show that SGPO-I is not necessarily monotonic: depending on the sampled traces and replay tasks, a round may produce steady gains, early saturation, temporary regression, or noisy recovery.
These patterns suggest that SGPO-I should be interpreted as a gated search process rather than a monotonic optimizer.
Temporary score drops are expected because each round samples new traces and regenerates rubrics and replay tasks.
The important property is not that every candidate update improves every replay set, but that paired replay exposes regressions before they are admitted into the production harness.

\subsubsection{Semantic-Gradient-based Prompt Optimization II}
\label{sec:harness:sgpo_ii}

As an extension of SGPO-I, SGPO-II changes the evidence source from dialogue traces to coding replay cases derived from historical merged requests (MRs).
This change is necessary because developing-agent failures are long-horizon and repository-specific: a generated patch may satisfy the surface requirement while violating ownership boundaries, feature-flag discipline, local helper conventions, rollback expectations, or deterministic validation rules.
The MR history therefore acts as the coding trace pool.
Before replay, SGPO-II filters this pool to remove noisy, stale, trivial, or bulk changes, including bulk reformats, reverts, dependency bumps, patches with fewer than 10 changed lines, rewrites above 5{,}000 changed lines, generated-file-heavy changes, and MRs whose original diff context no longer survives in the current codebase.
The retained examples are human-approved code changes with surviving repository context, suitable as evidence for coding-harness evolution.

\begin{table}[ht]
  \centering
  \caption{Five evaluation dimensions for coding patch replay.}
  \label{tab:coding_eval_dims}
  \resizebox{0.60\textwidth}{!}{%
  \begin{tabular}{lcp{5.8cm}}
    \toprule
    \textbf{Dimension} & \textbf{Weight} & \textbf{Criterion} \\
    \midrule
    Semantic Correctness   & 40\% & Core logic semantically equivalent to
                                    ground truth; boundary conditions correct
                                    (\emph{hard gate}: score $\geq 4$ required) \\
    Requirement Coverage   & 25\% & All acceptance criteria in the
                                    requirements specification satisfied \\
    File Coverage          & 20\% & Modified-file set consistent with the
                                    landed patch; no significant omissions \\
    Safety by Default      & 10\% & New behavior gated behind a feature flag
                                    disabled by default; no unguarded changes
                                    to existing logic \\
    Code Style Consistency &  5\% & Naming, structure, and comments follow
                                    repository conventions \\
    \bottomrule
  \end{tabular}
  }
\end{table}

From this cleaned pool, SGPO-II samples replay cases and converts each retained MR into a requirement-only task, analogous to the user tasks in SGPO-I.
The landed patch is hidden from the Developing Agent and used only by the evaluator as a reference after the agent has produced its implementation.
Cases are synthesized in 5 batch size before coding begins, and each case is attempted on a clean branch initialized from the MR base commit.
The rubric $\mathcal{R}$ is coding-specific: semantic correctness, requirement coverage, file coverage, safety, and style consistency (Table~\ref{tab:coding_eval_dims}).
The evaluator passes a case only when the weighted aggregate score is at least $4.0$ \emph{and} the Semantic Correctness dimension is at least $4$.
When replay fails, the evaluator summarizes the failure as a semantic gradient and passes it to a \textit{Harness Refine Agent} analogous to that of SGPO-I.In SGPO-II, the refinement agent converts that gradient into local updates to the Developing Agent's harness specification, specifically refining its execution rules, evaluator lessons, pattern memory, and deterministic prechecks.
Thus SGPO-II preserves the similar gradient-based optimization paradigm, while replacing session traces with repository-grounded coding evidence.

\begin{table}[ht]
\centering
\caption{SGPO-II case study: before and after self-evolution on a complex async module. The overall score improves from 2.60 to 4.90 ($+88\%$) after the harness accumulates project-specific constraints from earlier failures.}
\label{tab:sgpo_coding_case}
\resizebox{0.80\textwidth}{!}{%
\begin{tabular}{lp{0.38\textwidth}ccc}
\toprule
\textbf{Dimension} & \textbf{Failure description (Iteration 1)} & \textbf{Before} & \textbf{After} & \textbf{$\Delta$} \\
\midrule
Semantic Correctness   & Async callback coordination incomplete; approach diverges from reference & 2.5 & 5.0 & $+2.5$ \\
Requirement Coverage   & Critical async paths missing; fragmentary implementation & 2.6 & 4.9 & $+2.3$ \\
File Coverage          & Incomplete multi-file modifications; co-changes absent & 2.7 & 4.9 & $+2.2$ \\
Safety by Default      & Error handling incomplete across multiple paths & 2.5 & 5.0 & $+2.5$ \\
Code Style Consistency & Inconsistent formatting and naming conventions & 2.6 & 4.8 & $+2.2$ \\
\midrule
\textbf{Overall (weighted)} & & \textbf{2.60} & \textbf{4.90} & $\mathbf{+2.30}$ \\
\bottomrule
\end{tabular}%
}
\end{table}

\begin{figure}[ht]
  \centering
  \includegraphics[width=0.85\linewidth]{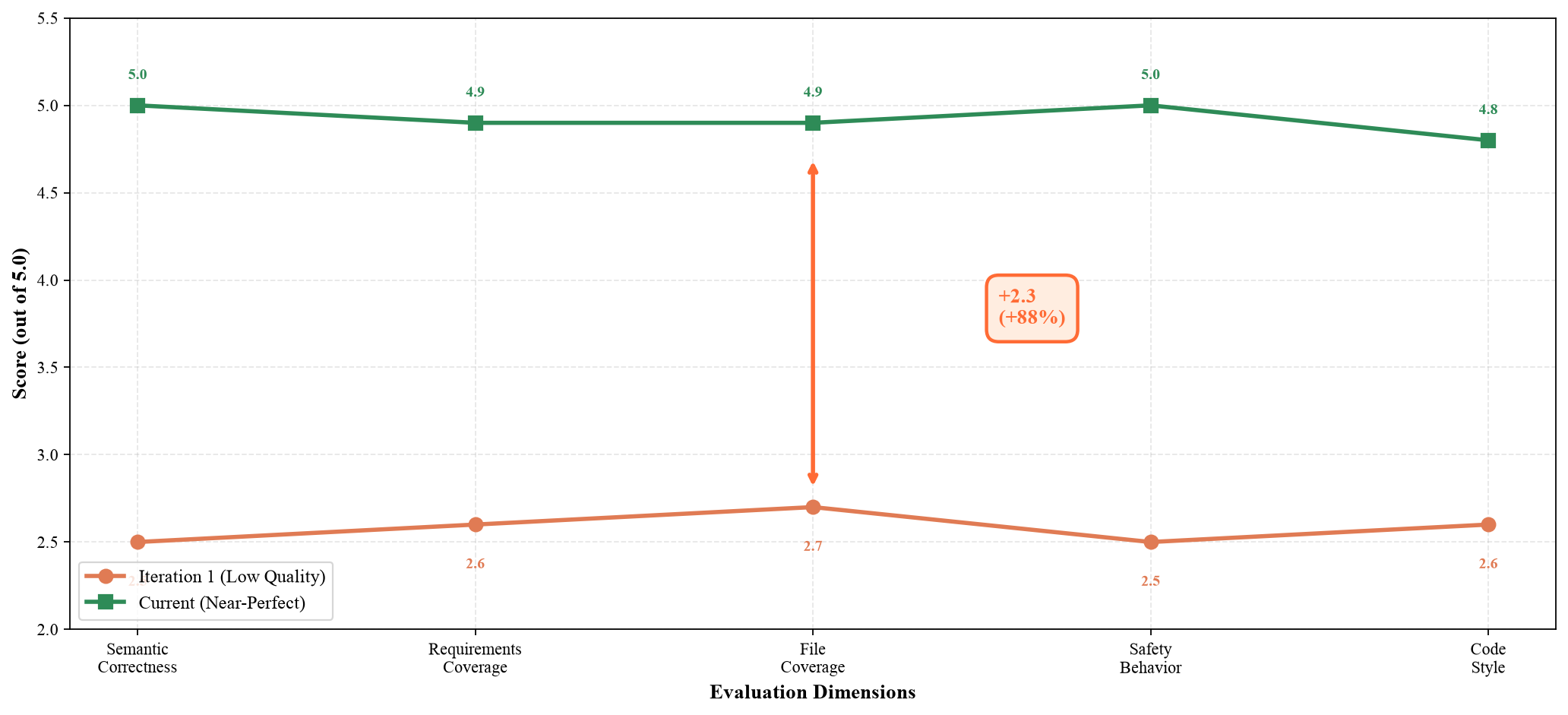}
  \caption{Case study: per-dimension score trajectory for a complex async module across SGPO-II self-evolution iterations. The harness converges from 2.60 to 4.90, with semantic correctness and safety reaching the maximum while style consistency shows the smallest remaining gap.}
  \label{fig:harness_case_study}
\end{figure}

In our production use case, SGPO-II is applied to a complex asynchronous module requiring multi-file coordination, async callback flow management, and comprehensive error handling.
Table~\ref{tab:sgpo_coding_case} summarizes the before-and-after snapshot; Figure~\ref{fig:harness_case_study} shows the per-dimension improvement trajectory across replay cases.

Notably, the iterative pipeline is not uniformly monotonic across all inputs.
We observe a small number of regressions in the evaluation set.
In one anonymized example, the score dropped from $3.67$ to $1.80$
($\Delta = -1.87$), resulting in a final \textsc{fail} judgment.
Such regressions arise under difficult or boundary-condition cases where the
LLM-as-judge misdiagnoses the failure mode and proposes a harness edit that
conflicts with an existing constraint.
We explicitly retain rather than filter these regressions: they are preserved
in the evaluation record to characterize method limits and guide targeted
follow-up improvements.
Critically, the paired-replay admission gate ($\Delta J > \epsilon$) prevents
regressive artifact updates from entering the accepted harness state.
A case that scores worse under the candidate harness than under the current
harness is rejected, so the judge's misdiagnosis produces a no-op rather than
a permanent degradation.
This makes the reliability of SGPO contingent on the admission gate rather
than on the infallibility of the evaluator.

\subsection{Evolving the Model-Research Exploration Pipeline}
\label{sec:evolve-model-track}

SGPO evolves \emph{how} the agents reason: it edits one subagent's harness---its instruction, validation rules, output contract, and tool discipline---and admits the edit only through paired replay.
A second axis of evolution concerns \emph{what} the system explores: the space of model-architecture directions itself.
Here the harness is fixed and the search space is what improves, as each experimental verdict reshapes which directions are generated next.
Given the populated facts store of Section~\ref{sec:research-agent}, the question is how to explore it systematically without relying on human curation of promising directions; the answer is a three-phase exhaustive loop, each phase building on conclusions from the previous.

\paragraph{Three-phase exploration.}
Phase~1 generates one reproduction proposal per paper, targeting its most cleanly evidenced finding.
These proposals are dispatched to the execution pipeline of Section~\ref{sec:developing-model-track} and the results aggregated into a leaderboard ranked by mean $\Delta$AUC against the production baseline.
The top $K{=}16$ papers advance to Phase~2.
This phase serves two purposes: it filters the search space to directions that actually transfer to the production setting, and it records, for each top-$K$ paper, which modules are marked \texttt{ablatable} in the architecture block.
Phase~2 isolates the contribution of each module.
For each ablatable module across the top-$K$ papers, the system enumerates two implementation choices---the original (\texttt{orig\_choice}) and an LLM-inferred alternative---and constructs bundles that activate only the target module while reverting all other top-$K$ additions to baseline equivalents.
The isolation is strict by design: bundled evaluations in the original papers often obscure which components actually drive the gain.
A module is confirmed effective if it produces positive $\Delta$AUC in at least 24 of the 32 resulting experiments.
Phase~3 then composes across papers.
Regular rounds extend each top-$K$ paper's lineage by grafting the highest-delta confirmed module not yet in its lineage, combining techniques that originated in different papers.
To prevent the search from converging prematurely on a small cluster of results, challenger rounds---triggered every four regular rounds---apply top effective modules to rank 17--32 papers, maintaining diversity in explored space.

\paragraph{Memory-guided pruning.}
A naive exhaustive search over 1{,}312 papers and their modules would be computationally prohibitive and would revisit directions already proven ineffective.
The system avoids this through a memory mechanism that continuously updates two levels of pruning decisions based on incoming experimental conclusions.
At the \emph{paper level}, if a result falsifies a paper's core premise in the production setting, all modules from that paper are excluded from Phases~2 and~3.
The RankUp paper provides a concrete example: a full reproduction run produced a best result of $\Delta\text{AUC}{=}{-}0.0203$, and training-log observables directly contradicted the paper's stated premise that effective-rank regularization improves representation diversity under online training.
Rather than proceeding to module ablation, the system recorded this as a paper-level prune and added an anti-pattern entry: verify that structural premises hold in the production setting before investing in module-level ablation.
At the \emph{module level}, Phase~2 ablations that confirm no independent contribution exclude the module from Phase~3 compositions.
Saturation detection provides a global stopping criterion: when the novelty-signature repeat rate---computed as a SHA-1 hash over discretized architecture fields across proposals---exceeds 80\% for two consecutive rounds, the loop terminates rather than generating structurally redundant experiments.
Together, phase-gating, memory-guided pruning, and saturation detection turn blind enumeration into a search that contracts around productive directions.

\paragraph{Experience flywheel.}
The conclusions that drive pruning are themselves curated rather than trusted on sight.
This is the offline-experiment counterpart of the negative-result assetization of Section~\ref{sec:analysis:negative_assetization}: where the Evaluation Agent turns discarded online A/B outcomes into reusable failure memory, the exploration loop turns every completed training round---successful or not---into a structured event appended to an append-only event log; the knowledge base is a view over this log that can be rebuilt from scratch at any time.
Lessons are organized into \texttt{anti\_patterns} (failure modes, each with log and diff regexes for recognizing analogous failures in future runs) and \texttt{playbook} (successful recipes, recorded when $\Delta\text{AUC} > 0.001$).
Not every experimental outcome is reliable enough to act on: a single successful run might reflect lucky hyperparameter interaction rather than a genuine mechanism, and a single failure might reflect a transient environment issue.
To filter signal from noise, each candidate experience entry must pass two gates before it is injected into future rounds.
The \emph{threshold gate} requires evidence from at least two independent runs.
The \emph{adversarial review gate} deploys a dedicated agent whose sole task is to falsify the claimed causal mechanism; entries that survive become \texttt{confirmed}, disputed entries become \texttt{contested} (injected with a warning), and falsified entries are permanently excluded.
This contrasts with the Skillhub in AgenticRecTune~\cite{agenticrectune}, which distills experience directly from A/B outcomes: an outcome alone does not explain why something worked, and an experience entry that cannot explain its own mechanism is a liability---it may generalize poorly, or encode a spurious correlation that poisons future decisions.
Requiring a verified causal explanation before promotion applies, at the knowledge-base level, the same standard that falsifiable attribution applies to a single round.

\paragraph{Closing the loop.}
These mechanisms compose into a self-evolving search.
Each experimental conclusion updates the exploration memory: confirmed playbook entries reinforce productive directions, anti-patterns trigger paper- or module-level pruning, and falsified premises close off entire branches of the search space.
The next round of exploration therefore starts from a search space that is not only larger---new papers arrive continuously through the daily ingest---but also sharper, annotated with the accumulated verdicts of every experiment the system has ever run.

\subsection{Discussion}
\label{sec:harness:discussion}

SGPO treats agent improvement as an offline, replayable system-optimization problem.
It complements online experiment analysis: experiment analysis explains whether a recommendation strategy worked, while SGPO explains how the agent harness should change when the trajectory itself reveals a recurring weakness.
The current implementation is intentionally conservative.
It updates one subagent specification at a time, freezes the rest of the operating harness, and admits edits only after paired replay.
This makes the update path slower than unconstrained self-editing, but safer for production recommendation workflows where a single overly broad harness change can corrupt downstream experiment artifacts.

\section{Experiments}
\label{sec:experiments}

\subsection{Production Measurement and Loop Performance}

To understand how agents reshape algorithm iteration, we study how AgentX
affects an engineer's iteration along three dimensions:
\begin{itemize}
  \item \textbf{RQ1.} Does AgentX shorten the idea-to-rollout cycle?
  \item \textbf{RQ2.} Does AgentX deliver more rollout-level results per worker?
  \item \textbf{RQ3.} Does the experiments that AgentX rolls out
    deliver measurable online metric gains?
\end{itemize}
Answering these questions requires visibility into the full lifecycle. We therefore build a monitoring platform that instruments every stage, and analyze the conversion metrics it produces.

\paragraph{Monitoring and Data Validity.}
For longitudinal evaluation of the AgentX auto-iteration pipeline, we
instrument the experiment lifecycle across all workers: experiment documents,
review states, coding and deployment events, and AB decisions are normalized
into an append-only event log aligned to canonical lifecycle stages. Adjacent
conversion rates are computed under a fixed snapshot scope, so any rerun over
the same window reproduces the same aggregates and stage attribution. The
resulting node counts and adjacent conversion rates form the chained funnel
reported in Table~\ref{tab:node_rate} and Eq.~\ref{eq:funnel}.

We answer the \textbf{three} research questions with a single
body of evidence collected from a \textbf{three-week} deployment in which
\textbf{three AgentX} workers ran idea-to-rollout loops concurrently across two
production scenarios on the \textbf{Kuaishou App}: main feed
recommendation and life-service commercialization. The monitoring platform logs every node of the loop---idea
pass, code-and-launch, and positive evaluation---as an explicit state
transition, where a \emph{positive evaluation} denotes an experiment that
qualifies for full rollout, i.e., a launchable result (LR).
Table~\ref{tab:node_rate} reports the per-node conversion rates, and
Table~\ref{tab:human_vs_agent} contrasts AgentX with a single algorithm
engineer on per-worker productivity.

\begin{table}[htbp]
  \centering
  \caption{Node-level conversion rates and counts of the AgentX idea-to-rollout iteration loop.}
  \label{tab:node_rate}
  \begin{tabular}{lccccc}
    \toprule
    \textbf{Scenarios} & \textbf{Ideas} & \textbf{Idea Pass} & \textbf{Code-\&-Launch} & \textbf{Positive Eval.} & \textbf{LR} \\
    \midrule
    Main Feed & 361 & 27.7\% & 95.0\% & 8.4\%  & 8  \\
    Life Service        & 13  & 46.1\% & 83.3\% & 40.0\% & 2  \\
    \textbf{Overall}  & \textbf{374} & \textbf{28.34\%} & \textbf{94.3\%} & \textbf{9.9\%} & \textbf{10} \\
    \bottomrule
  \end{tabular}
\end{table}

\noindent
The pipeline forms a chained conversion funnel, where each stage is conditioned
on the success of the previous one:
\begin{equation}
  \label{eq:funnel}
  \underbrace{374}_{\text{ideas}}
  \xrightarrow{\text{idea pass } 28.34\%}
  \underbrace{106}_{\text{passed}}
  \xrightarrow{\text{code\&launch } 94.3\%}
  \underbrace{100}_{\text{launched}}
  \xrightarrow{\text{positive eval. } 9.9\%}
  \underbrace{10}_{\text{LR}}.
\end{equation}
The two business lines are internally consistent under this formulation:
$361 \times 27.7\% \times 95.0\% \times 8.4\% = 8$ (Main Feed) and
$13 \times 46.1\% \times 83.3\% \times 40.0\% = 2$ (Life Service).

\begin{table}[t!]
  \centering
  \caption{Categorized rejection reasons for the 268 ideas that failed the
    \textsc{validate} gate. Each idea is assigned to exactly one
    category by priority $A \to B \to C \to E \to D \to F \to Z$.}
    \label{tab:reject_reasons}
  \small
  \begin{tabular}{clr}
    \toprule
    \textbf{ID} & \textbf{Rejection Category} & \textbf{Ratio} \\
    \midrule
    A & Parameter-resource conflict       & 64.7\% \\
    B & Pre-condition \texttt{enable} flag off        &  6.4\% \\
    C & Violates hard constraint, whitelist, or capability boundary                     &  7.5\% \\
    E & Infeasible implementation / parameter not found / missing prerequisite code     &  0.6\% \\
    D & Overlap with historical experiments                              &  5.2\% \\
    F & Weak hypothesis / guardrail risk / null-result risk                             &  1.2\% \\
    Z & Missing user/item attribute                                    & 14.5\% \\
    \textbf{Total} & & \textbf{100.0\%} \\
    \bottomrule
  \end{tabular}
\end{table}

\begin{table}[t!]
  \centering
  \caption{Categorized failure modes observed in the \textsc{coding} phase.}
  \label{tab:coding_fail_reasons}
  \small
  \begin{tabular}{clr}
    \toprule
    \textbf{ID} & \textbf{Coding Failure Category} & \textbf{Ratio} \\
    \midrule
    CA & DSL / force-enable wiring error on the debug branch                       & 35\% \\
    CB & C++ / MaTX compiler constraint violation                                  & 20\% \\
    CC & Mis-structured \texttt{if/else} branches across parallel ideas            & 15\% \\
    CD & Dryrun state-machine misjudgment                                          & 10\% \\
    CE & Local validation environment unavailable                                  & 10\% \\
    CF & \texttt{dryrun-mr} launch-contract or dirty-worktree violation            &  5\% \\
    CG & Log-parsing artefact masquerading as a code bug                           &  5\% \\
    \textbf{Total} & & \textbf{100\%} \\
    \bottomrule
  \end{tabular}
\end{table}

\noindent
\textbf{What blocks an idea before coding and launch?}
Table~\ref{tab:reject_reasons} partitions the 268 rejected ideas into a
mutually exclusive taxonomy and exposes a striking asymmetry: the failures
that look like the agent's responsibility are in fact concentrated outside it.
\emph{Parameter-resource conflict} at the AB platform alone accounts for
\textbf{64.7\%} of all rejections (A)---the target parameter is already
held by a holdout combo experiment, an in-flight experiment, or a different traffic
world, so the idea is technically sound yet cannot be materialized on
production traffic. Three adjacent operational frictions contribute another
\textbf{12.2\%}: soft-switch oversights in which the prerequisite
\texttt{enable} flag is still off (B, 6.4\%), duplicates of historical
launchable results or in-flight experiments that would only dilute existing
traffic (D, 5.2\%), and infeasible implementations whose target parameter is
absent from the live DSL world (E, 0.6\%). A further \textbf{14.5\%} (Z)
falls on the data-infrastructure side: the idea relies on a user- or
item-side signal that is not yet exposed in the production feature store,
so the experiment cannot be wired up no matter how sharp the agent's
reasoning. Taken together, platform and infrastructure constraints absorb
\textbf{91.4\%} of all rejections, while genuine agent errors---hard-constraint
or whitelist violations (C, \textbf{7.5\%}) and weak hypotheses or guardrail
risks (F, \textbf{1.2\%})---make up only \textbf{8.7\%}. Two implications
follow. The bottleneck of the loop is operational rather than
algorithmic---closing the A-class gap alone would recover roughly two thirds
of the currently lost ideas; and the highest-leverage next step is therefore
not a smarter agent but an \emph{upstream conflict checker} that queries the
AB-platform state \emph{before} brainstorm, so that resource-locked
parameters never enter the candidate pool in the first place.

\noindent
\textbf{Where do the residual coding loop failures concentrate?}
Table~\ref{tab:coding_fail_reasons} exposes the same asymmetry as
Table~\ref{tab:reject_reasons}: over \textbf{95\%} of failures are
infrastructure-side rather than agent-side. Framework-syntax violations on the debug branch---DSL wiring
(CA, 35\%), MaTX/C++ constraints (CB, 20\%), and \texttt{if/else} structure
(CC, 15\%)---absorb \textbf{70\%} of all coding failures. Toolchain and
environment friction (CD+CE) add another \textbf{20\%}. Genuine algorithmic
mistakes by the coding agent account for under \textbf{5\%}. The takeaway
parallels Table~\ref{tab:reject_reasons}: the residual losses trace to
framework conventions that could be front-loaded as templates, not to the
agent's reasoning quality.

\paragraph{AgentX shortens the idea-to-rollout cycle by turning a serial
workflow into a parallel pipeline(RQ1).}
A manual workflow executes the lifecycle as a single serial chain: each idea
blocks the next and throughput is capped by the slowest hand-off. AgentX
restructures the chain into a \emph{parallel pipeline} by decoupling the
proposal, coding, launch, and monitoring stages into independent workers, so
distinct ideas occupy different stages at the same instant. The effective
cycle time per idea therefore collapses to that of the busiest stage rather
than the sum of all stages. Empirically, the three workers carried
\textbf{374} ideas through the loop in three weeks, and per-worker concurrent
throughput roughly doubled each week as the system self-evolved through
skill consolidation, pitfalls accumulation, and dryrun-template maturation.
Averaged across the three-week window, each AgentX worker sustained roughly
\textbf{12} concurrent experiments, against \textbf{1.5} for an engineer
under the traditional manual workflow---an \textbf{8$\times$} single-worker
gain (Table~\ref{tab:human_vs_agent}).
Because no stage
forces ideas to contend for a shared resource, adding workers raises throughput
almost linearly, so iteration speed is governed by idea supply and evaluation
bandwidth rather than by how fast one person can push a single idea through
the chain.

\paragraph{AgentX delivers more rollout-level results by
automating both ends of the loop under human review(RQ2).}
At the front end, a \emph{brainstorm} agent produces candidate ideas in
parallel and routes them through a review gate, so engineers curate a
high-volume stream instead of hand-writing each idea; only the \textbf{28.34\%}
of ideas that pass review survive (Table~\ref{tab:node_rate}). Human effort
is thereby elevated from idea production to idea selection. At the back end,
the \emph{developing} agent is the decisive lever: turning an idea into deployable
code that clears dry-run validation is the most labor-intensive and
failure-prone step of a manual workflow. By raising the code-and-launch rate
to \textbf{94.3\%}, the developing agent absorbs nearly all of this work, loses
few ideas, requires no human attendance, and lets engineers focus on review
rather than execution. Over the three-week window the three workers converted
374 ideas into 10 launchable results (end-to-end conversion
$10/374 \approx 2.67\%$), or roughly \textbf{3.3} LRs per worker. We do not
claim parity on per-idea quality: as Table~\ref{tab:human_vs_agent} shows, a
senior engineer still wins on per-idea hit rate by about
\textbf{1.9$\times$}, because human-authored ideas are hand-picked and
pre-filtered. AgentX deliberately trades \emph{precision under scarcity} for
\emph{volume under automation}: at the worker-week granularity it produces
\textbf{0.0623\%} cumulative app-time gain against \textbf{0.0167\%} for an
engineer---a \textbf{3.7$\times$} lift in realized business value per unit of
human capacity. Automating both ends of the loop is therefore a concrete step
toward fully automating the algorithm-iteration workflow.

\begin{table}[htbp]
  \centering
  \caption{Per-worker productivity and concurrency: AgentX worker vs algorithm engineer under the traditional manual-iteration workflow.}
  \label{tab:human_vs_agent}
  \begin{tabular}{lccc}
    \toprule
    \textbf{Metric (per worker $\cdot$ week)} & \textbf{AgentX} & \textbf{Engineer} & \textbf{Ratio} \\
    \midrule
    Concurrent experiments & 12        & 1.5      & \textbf{8$\times$} \\
    LR Count
       & 1.1       &0.08      &\textbf{13.8$\times$} \\
    Cumulative app-time gain produced    & 0.0623\%  & 0.0167\% & \textbf{3.7$\times$} \\
    Per-idea rollout conversion rate     & 2.7\%     & 5.1\%   & \textbf{0.53$\times$} \\
    \bottomrule
  \end{tabular}     
\end{table}

\begin{figure}[ht]
  \centering
  \includegraphics[width=0.75\linewidth]{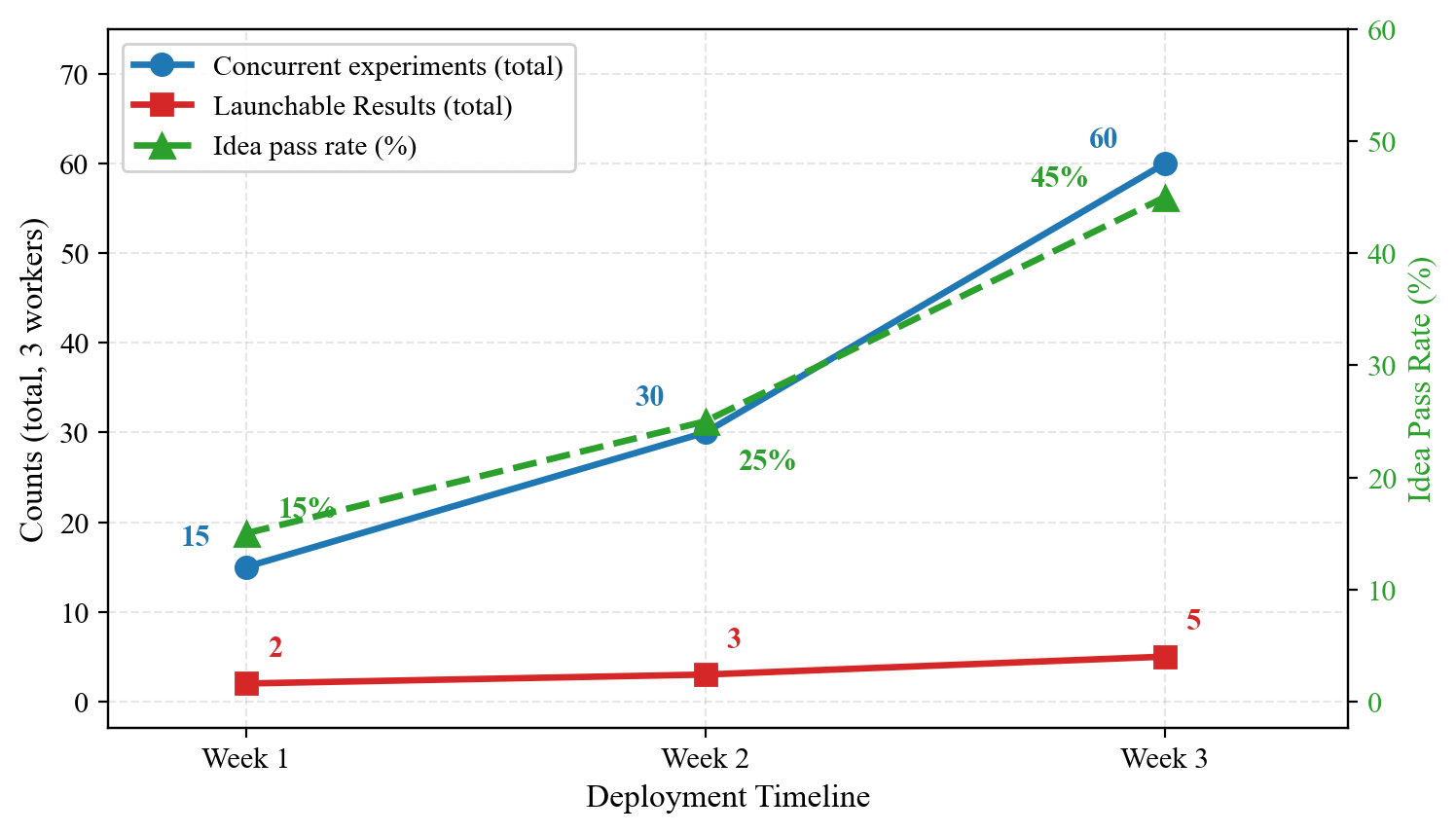}
  \caption{Three weeks of self-evolution transformed the AgentX loop:
    weekly concurrent experiments \textbf{quadrupled} (15$\to$60), idea pass
    rate \textbf{tripled} (15\%$\to$45\%), and weekly launchable results
    \textbf{more than doubled} (2$\to$5). Skill consolidation, pitfalls
    accumulation, and dryrun-template maturation compounded throughput with
    selectivity---the system no longer just produces more, it produces more
    of the right ideas.}
  \label{fig:weekly_ramp}
\end{figure}

\paragraph{The experiments AgentX rolls out deliver measurable online
gains, and their volume scales with compute(RQ3).}
The system produced 10 launchable results end to end---8 from the 361
main feed ideas and 2 from the 13 life-service ideas---and these rollouts
translated into sizable business gains:
\begin{itemize}
  \item \textbf{Main Feed.} A cumulative gain of \textbf{0.561\%} in user
    app consumption time on Kuaishou App.
  \item \textbf{Life Service.} Over \textbf{RMB 100 million} in annualized
    revenue for the Kuaishou platform.
\end{itemize}
As Figure~\ref{fig:weekly_ramp} visualizes, self-evolution simultaneously
expanded throughput and tightened selectivity over the three-week window. Because human effort is confined to review and decision-making while every
execution-heavy stage is automated, the volume of rollout-level results is no
longer bounded by headcount but primarily by the compute allocated to the
loop. Throughput scales roughly linearly with the number of workers---each
additional worker contributes LRs at the ~1.1-per-worker-per-week rate
observed here---and during the ramp phase the system's self-evolution
compounded this with an additional \textbf{doubling} per week. The business value
demonstrated in this study can therefore be expanded along two complementary
axes: more workers (linear in compute) and a more mature system
(super-linear early on).

\subsection{Model Iteration Funnel of AgentX}
\label{sec:experiments:model_funnel}
We conduct a standalone study on agentX-driven recommendation model research and development. 
This study focuses on using AgentX for an end-to-end model iteration workflow, including autonomous paper reproduction, module ablation, cross-paper architectural composition, and new architecture discovery. 
Importantly, this study is \emph{not} part of the three-week production deployment summarized in Table~\ref{tab:node_rate}; its node counts and conversion rates are computed on a separate experiment pool and are therefore \emph{not directly comparable} to Table~\ref{tab:node_rate}.

\paragraph{Study setup.}
We curate a candidate pool of recent recommendation papers and ask AgentX to reproduce each paper on top of a unified codebase, following the same protocol used in Appendix~\ref{sec:detailed_experiment}. 
We then rank the reproduced models by their aggregated performance across four public datasets: KuaiRand~\cite{kuairand}, Taobao~\cite{taobao}, Amazon~\cite{amazon}, and ML-1M~\cite{movielens}. 
The top-performing reproduced models are selected as the seed set for cross-paper architectural composition.
For each seed paper, AgentX automatically identifies complementary modules from the other seed papers that are not implemented in the current seed model. 
It then synthesizes these modules into the seed architecture through LLM-driven code composition. 
Each composed model is registered as a new experiment and re-enters the same leaderboard, allowing successful compositions to become candidate seeds for further rounds of iteration.

Throughout this study, the comparison baseline is the production ranking algorithm currently deployed on Kuaishou App, simplified only to fit the public-dataset feature schema. 
This simplification removes business-only features and serving constraints while preserving the core architecture and training recipe. 
Therefore, an experiment that ``beats baseline'' in this setting should be interpreted as a necessary, but not sufficient, indicator of potential business value.

\paragraph{Funnel.}

The model-agent loop follows a two-stage funnel. 
The first stage is a \emph{selection} stage, where AgentX distills a large pool of reproduced papers into a smaller high-quality seed set. 
The second stage is an \emph{expansion-and-validation} stage, where AgentX automatically dispatches cross-paper architectural compositions and filters them through offline evaluation, online testing, and launch-review gates.
Detailed results are presented in Table~\ref{tab:model_node_rate}.

\begin{table}[htbp]
  \centering
  \caption{Node-level counts and adjacent conversion rates of the
    \emph{model-agent} iteration funnel.
    \textbf{This is a dedicated standalone study and is not part of the
    three-week production deployment reported in Table~\ref{tab:node_rate}};
    the two funnels are computed on disjoint experiment pools and are not
    directly comparable. The baseline is the production ranking algorithm of
    Kuaishou App, simplified only to fit the public-dataset feature schema.}
  \label{tab:model_node_rate}
  \small
  \begin{tabular}{lcc}
    \toprule
    \textbf{Stage} & \textbf{Count} & \textbf{Adj.\ rate} \\
    \midrule
    Candidate papers (reproduced)          & 113 & ---     \\
    Auto-dispatched experiments            & 214 & --- \\
    Completed runs                         & 180 & 84.1\%  \\
    Beats baseline (public datasets)       & 104 & 57.8\%  \\
    Sent to Online A/B            & 22  & 21.2\%  \\
    Online A/B positive                        & 5   & 22.7\%  \\
    Launchable result (LR)                 & 2   & 40.0\%  \\
    \bottomrule
  \end{tabular}
\end{table}

\paragraph{Findings.}
Three observations distinguish the model funnel from the strategy funnel reported in Table~\ref{tab:node_rate}. 
First, the \textbf{expansion-then-pruning} shape reflects how AgentX uses offline experimentation to reduce the cost of online exploration. 
Starting from the seed models, AgentX first expands the search space into 214 cross-paper architectural compositions. 
It then filters out candidates with insufficient potential through run-completion checks, offline benchmark performance, and business-side dispatch criteria. 
As a result, only 22 experiments are promoted to online A/B testing, reducing the online testing load by nearly an order of magnitude before consuming production traffic.

Second, the \textbf{$\mathbf{57.8\%}$ beats-baseline rate} should be interpreted together with the baseline simplification described in the setup. Since the public-dataset baseline removes business-only features and serving constraints that the production algorithm relies on, this rate does not mean that the candidate models beat the full production system as deployed. Instead, it indicates that the candidate is a legitimate module or architecture worth considering for production integration.

Third, the \textbf{22 $\to$ 5 $\to$ 2} contraction on the right side of the funnel is the tightest section. Even after offline gains are observed, the joint requirement of online-positive movement and launch-review approval reduces the candidate set by roughly an order of magnitude. This pattern is consistent with the 9.9\% positive-evaluation rate in the production funnel, suggesting that the online-review gate, rather than offline accuracy or coding throughput, is the dominant bottleneck for model-side LR yield.

\paragraph{Takeaway.}
This standalone model-agent study shows that the model-iteration component of AgentX can close an end-to-end loop from paper reproduction and architectural composition to offline evaluation, online testing, and launchable results. 
On a disjoint experiment surface, the system produces LR-grade artifacts from both the original paper pool and the auto-dispatched experiment pool: 2 out of 113 candidate papers and 2 out of 214 composed experiments reach LR. The two LR models yield a \textbf{0.865\%} gain in live streaming durationon Kuaishou App.

Together with the production funnel in Table~\ref{tab:node_rate}, this result suggests that AgentX is not specialized to a single research surface. Instead, it instantiates a portable closed-loop pattern that can be applied to both strategy-side and model-side recommendation system iteration. Detailed per-method results, module-combination studies, and the industrial-scenario transfer used in this loop are provided in Appendix~\ref{sec:detailed_experiment}.

\subsection{Showcase I: End-to-end AgentX Experiments}
\label{sec:experiments:end_to_end_showcase}

The strongest showcase for the current AgentX loop is not a single isolated metric, but the fact that multiple launch-review documents can be traced back to an autonomous research path: proposal generation, feasibility checking, implementation, online experimentation, and post-experiment analysis. This demonstrates that AgentX is already capable of completing the end-to-end cycle in realistic recommendation scenarios.

\paragraph{PCV-enhanced constrained fine-ranking score.}
This launch-review case demonstrates how AgentX improves a mechanism-level ranking idea through a closed multi-agent loop.
The primary goal was to increase user watch time in the main Kuaishou feed while keeping user real-show stable.
To achieve this goal, AgentX introduced post-consumption value (PCV) as an additional ranking signal.
PCV captures behaviors after video consumption, such as sharing, collecting, and replaying, and can indicate lasting content value beyond immediate watch-time or click signals.
However, PCV is also risky: high PCV does not always imply high content quality, since low-quality viral or bait-like content may also trigger post-consumption behaviors.
We next describe the end-to-end AgentX experimentation process for this case.

\begin{table*}[ht]
\centering
\caption{Loop 1: Direct PCV boosting -- agent inputs and outputs.}
\label{tab:loop1_agents}
\small
\begin{tabular}{p{2.5cm}p{4.2cm}p{8.8cm}}
\toprule
\textbf{Agent} & \textbf{Input} & \textbf{Output} \\
\midrule
\textbf{Brainstorm Agent}
& Ranking objective: improve user watch time while keeping real-show and user-experience guardrails stable.
& Generated five candidate ideas:
(1) session-budget-aware scoring,
(2) duration-preference matching,
(3) exploration-exploitation balancing,
(4) negative-feedback-sensitive user protection,
and (5) post-consumption-value boosting.
The agent selected post-consumption-value boosting as the launch candidate, based on the hypothesis that existing fine-ranking scores under-utilize post-consumption behaviors that reflect long-tail content value. \\
\midrule
\textbf{Developing Agent}
& Selected PCV boosting idea.
& Implemented a production-feasible multiplicative formula:
$S_1 = B_r \cdot (1 + \beta P)$,
guarded by experiment switches and ready for online A/B evaluation. \\
\midrule
\textbf{Evaluation Agent}
& Online A/B result of $S_1$.
& Produced the first-round diagnosis: weak positive but statistically unreliable duration gains
($+0.034\%$ per-capita watch time; $+0.021\%$ user watch time), with diagnostic risks such as active devices $-0.023\%$, and the 18–30 age segment in device-average usage $-0.032\%$.
Conclusion: direct PCV boosting was too noisy and needed quality and duration constraints. \\
\bottomrule
\end{tabular}
\end{table*}

\begin{table*}[ht]
\centering
\caption{Loop 2: Constrained PCV ranking -- agent inputs and outputs.}
\label{tab:loop2_agents}
\small
\begin{tabular}{p{2.5cm}p{4.2cm}p{8.8cm}}
\toprule
\textbf{Agent} & \textbf{Input} & \textbf{Output} \\
\midrule
\textbf{Brainstorm Agent}
& Loop-1 evaluation feedback: direct PCV boosting was directionally positive but not sufficiently robust.
& Refined the idea into constrained PCV ranking with three design principles:
quality gating, activity-aware dynamic weighting, and a duration-oriented base score. \\
\midrule
\textbf{Developing Agent}
& Constrained PCV ranking idea.
& Implemented the constrained scoring mechanism:
$S_2 = B_d \cdot (1 + \beta(u)G(P))$,
and prepared the second-round online A/B test. \\
\midrule
\textbf{Evaluation Agent}
& Online A/B result of $S_2$.
& Produced a launch-review-ready conclusion and consolidated the two-loop learning into a reusable knowledge artifact: direct PCV boosting is noisy, while constrained PCV with quality gating, activity-aware weighting, and a duration-oriented base score yields clearer gains.
The final online result showed user watch time $+0.071\%$ and real-show $+0.118\%$, while keeping user-experience guardrails stable. \\
\bottomrule
\end{tabular}
\end{table*}

\textbf{Loop 1: Direct PCV boosting.}
As summarized in Table~\ref{tab:loop1_agents}, the first loop tested whether PCV could be directly used as an additional ranking signal.
The Brainstorm Agent generated several candidate directions and selected PCV boosting.
The Developing Agent converted the idea into a simple multiplicative scoring formula:
\begin{equation}
S_1 = B_r \cdot (1 + \beta P),
\end{equation}
where $B_r$ denotes the relevance-oriented base score, $P$ denotes the mixed PCV score, and $\beta$ is a fixed PCV weight.
The Evaluation Agent then analyzed the online A/B result.
The first version showed positive but weak gains on the main duration metrics: per-capita watch time improved by $+0.034\%$ , and user watch time improved by $+0.021\%$ .
However, several diagnostic metrics suggested potential instability: active devices moved slightly negative ($-0.023\%$), and the 18--30 age segment showed negative movement in device-average usage ($-0.032\%$).
Diversity-related metrics also showed small negative movements, such as effective interest counts, novelty clusters, and surprise clusters.

The Evaluation Agent therefore diagnosed that direct PCV boosting was directionally promising but too noisy.
The likely reason is that the first version boosted all high-PCV content without distinguishing high-quality post-consumption value from noisy or bait-like post-consumption behaviors.
It also used a relevance-oriented base score and a fixed PCV weight, which provided no explicit protection for the watch-time objective and no adaptation to user activity level.
This diagnosis became the input to the second loop: PCV should be introduced only under explicit quality gating, activity-aware weighting, and duration-oriented constraints.

\textbf{Loop 2: Constrained PCV ranking.}
As summarized in Table~\ref{tab:loop2_agents}, the second loop used the first-round evaluation as input.
The Brainstorm Agent refined the idea from direct PCV boosting to constrained PCV ranking.
The Developing Agent implemented the constrained formula:
\begin{equation}
S_2 = B_d \cdot (1 + \beta(u)G(P)),
\end{equation}
where $B_d$ denotes the duration-oriented base score, $G(P)=\max(P-\tau,0)$ is the quality-gated PCV signal with threshold $\tau$, and $\beta(u)$ is an activity-aware dynamic weight determined by user activity level.
The Evaluation Agent then validated the second-round design through online A/B testing.
The constrained PCV mechanism achieved clearer positive lifts: user watch time increased by $+0.071\%$, and real-show increased by $+0.118\%$, while user-experience guardrails remained stable.
Beyond metric validation, the Evaluation Agent also consolidated the two-loop learning into a reusable knowledge artifact: direct PCV boosting can be noisy, while constrained PCV ranking becomes more reliable when PCV is filtered by quality, scaled by user activity, and anchored to a duration-oriented base score.

This case demonstrates that AgentX is not a one-shot idea generator.
It closes the loop from idea generation to implementation, online evaluation, feedback-driven redesign, knowledge consolidation, and measurable online impact.

\subsection{Showcase II: Co-evolution with Expert Agents}
\label{sec:experiments:agent_for_rec_showcase}

Beyond the standalone end-to-end mode, we further deploy a multi-agent collaboration mode in which an expert agent operates in concert with AgentX in the life-service scenario. Concretely, we train a dedicated recommendation decision agent as the expert agent for this scenario. It performs user-level diagnosis by observing user behavior and profiles, assessing whether the current exposure aligns with a user's deeper service interests, and producing control decisions expressed in natural language. AgentX then reasons about how each diagnosis can be addressed and whether it is feasible in production, refines it into a sounder scheme, and writes a complete, deployable execution plan that it ships online.

\begin{figure}[ht]
  \centering
  \includegraphics[width=0.8\linewidth]{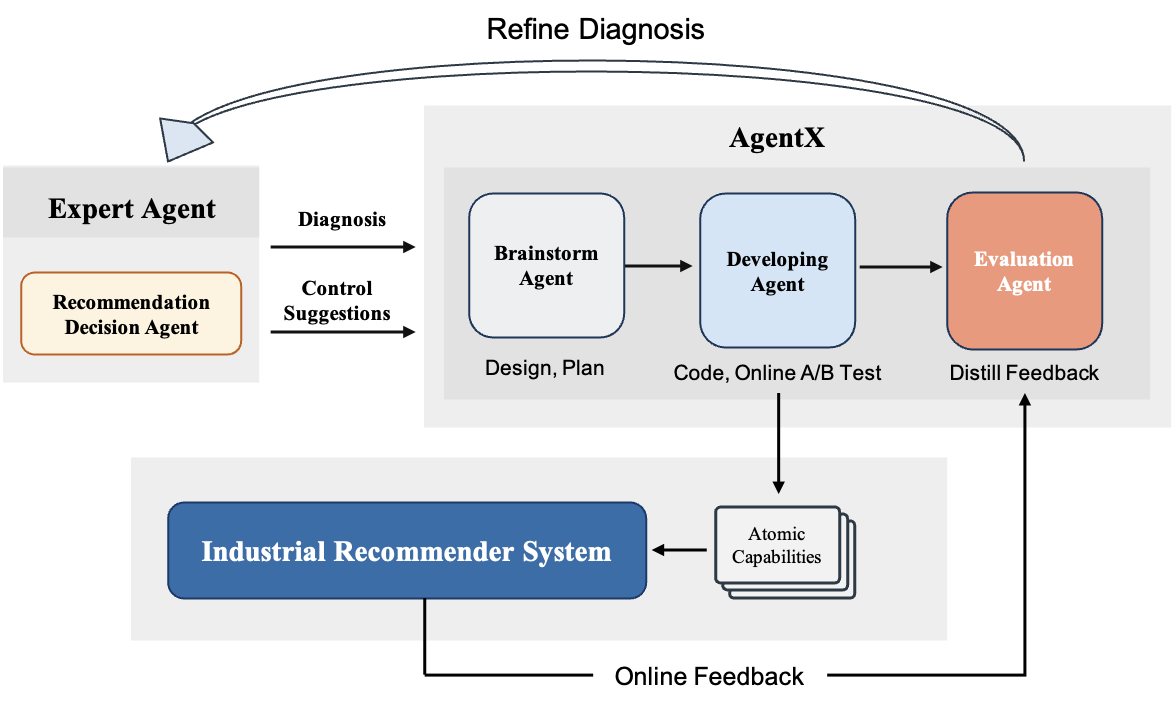}
  \caption{Production flow for the life-service recommendation decision agent. AgentX converts diagnostic signals into production operators, while the resulting operators expand the atomic action space available to the decision agent.}
  \label{fig:agentx_reco_decision_flow}
\end{figure}

\begin{table}[t!]
\centering
\caption{Expert agent and AgentX co-evolution loop -- agent inputs and outputs. \textbf{Revenue} refers to the commercial revenue generated in the online advertising system of the Kuaishou platform.}
\label{tab:locallife_agents}
\small
\begin{tabularx}{\textwidth}{p{3.0cm}p{3.8cm}X}
\toprule
\textbf{Agent} & \textbf{Input} & \textbf{Output} \\
\midrule
\textbf{Recommendation Decision Agent} (Expert)
& History sequence, user profile, and aggregated statistics.
& Diagnosed the user's advertising interests via chain-of-thought reasoning—e.g., combining a driver occupation with searches for vehicle pricing to infer a clear but under-served interest in automotive ads—and emitted control suggestions across multiple factors (industry filter, CPM threshold, CPM boost ratios, etc.) as structured JSON. \\
\midrule
\textbf{Brainstorm Agent}
& Control suggestions from the recommendation decision agent.
& Assessed the feasibility of the control suggestions and selected a CPM boost as the control action, producing a complete, executable scheme for it. \\
\midrule
\textbf{Developing Agent}
& The CPM-boost scheme.
& Implemented the data-fetching logic and the UV-level control, carrying the scheme through code development to an online A/B test. \\
\midrule
\textbf{Evaluation Agent}
& Online A/B result of the CPM-boost scheme.
& Produced a launch-review-ready conclusion: the UV-level CPM boost achieved a positive lift, increasing revenue by $+4.7\%$. \\
\bottomrule
\end{tabularx}
\end{table}

Several sub-agents within AgentX jointly support this deployment pipeline. The Brainstorm Agent assesses the feasibility of the control suggestions produced by the recommendation decision agent, designs the concrete atomic controls, and settles on a final implementation scheme. The Developing Agent then realizes the scheme as a controllable online action, carrying it through code development to deployment. The Evaluation Agent reads online feedback and distills the results into the next iteration of the decision policy. Figure~\ref{fig:agentx_reco_decision_flow} provides an overview of this multi-agent collaboration pipeline, from the expert agent's diagnosis to AgentX's deployment and the online feedback loop. Table~\ref{tab:locallife_agents} summarizes the inputs and outputs of each agent in one such loop.

This co-evolution has already demonstrated its value in production. In one life-service record, the recommendation decision agent achieved a relative revenue lift of $+4.7\%$ on a large user cohort. These results indicate that the value of AgentX lies not only in autonomously completing end-to-end research, but also in serving as an amplifier for the entire recommendation pipeline. For a request of any form, it rapidly reasons out an executable scheme, carries it through to full deployment, and continuously monitors it. In this process, the recommendation decision agent and AgentX co-evolve along two complementary dimensions, where the recommendation decision agent, fine-tuned on life-service domain knowledge, serves as a domain expert that supplies high-quality diagnosis and deepens the research loop, while AgentX continuously proposes, implements, and validates new strategy directions and realizes these diagnoses as fine-grained atomic actions, thereby broadening the loop. High-quality execution traces further refine AgentX through SGPO, yielding cumulative capability gains across iterations.

\section{Conclusion}\label{sec:conclusion}

In this technical report, we have presented \textbf{AgentX}, an automated development framework that drives the end-to-end iteration of industrial recommender systems through a closed loop of agents. AgentX decomposes the production lifecycle into role-specialized modules, including a Brainstorm Agent that produces evidence-grounded experiment proposals, a Developing Agent that translates them into production-ready code, and an Evaluation Agent that manages safe rollout and online A/B judgment. Wrapping these modules, a trajectory-driven harness evolution layer distills execution traces into updates of agent prompts, workflows, and the underlying foundation model, so that the system grows stronger with each round of iteration rather than resetting to zero. Grounded in real online A/B feedback rather than offline metrics, AgentX has delivered an order-of-magnitude improvement in both experiment throughput and cumulative online gain in a three-week production window.

Beyond these concrete results, the deeper significance of AgentX lies in reshaping the production function of recommender system iteration. The throughput of recommendation automation has long been constrained by a structural fact: high-quality ideas hinge on the manpower and inspiration of algorithm engineers, which is linear, scarce, and hard to replicate. Once agents take over idea generation, code landing, and preliminary validation at scale, the role of engineers shifts from personally executing every experiment toward defining which problems are worth exploring and designing more efficient agent systems. Trajectory data connects this shift, since every improvement to the underlying system is automatically inherited by every subsequent experiment. The iteration of industrial recommender systems is thereby rewritten from a linear addition of human labor into a compoundable, engineerable leverage. We release this report to invite the broader community to explore agent-driven iteration in their own systems.

\printbibliography

@article{onerec,
  title={Onerec: Unifying retrieve and rank with generative recommender and iterative preference alignment},
  author={Deng, Jiaxin and Wang, Shiyao and Cai, Kuo and Ren, Lejian and Hu, Qigen and Ding, Weifeng and Luo, Qiang and Zhou, Guorui},
  journal={arXiv preprint arXiv:2502.18965},
  year={2025}
}

@article{onelive,
  title={OneLive: Dynamically Unified Generative Framework for Live-Streaming Recommendation},
  author={Wang, Shen and Huang, Yusheng and Yang, Ruochen and Wen, Shuang and Xu, Pengbo and Cao, Jiangxia and Liu, Yueyang and Cai, Kuo and Guo, Chengcheng and Wang, Shiyao and others},
  journal={arXiv preprint arXiv:2602.08612},
  year={2026}
}

@inproceedings{din,
  title={Deep interest network for click-through rate prediction},
  author={Zhou, Guorui and Zhu, Xiaoqiang and Song, Chenru and Fan, Ying and Zhu, Han and Ma, Xiao and Yan, Yanghui and Jin, Junqi and Li, Han and Gai, Kun},
  booktitle={Proceedings of the 24th ACM SIGKDD international conference on knowledge discovery \& data mining},
  pages={1059--1068},
  year={2018}
}

@inproceedings{rankmixer,
  title={Rankmixer: Scaling up ranking models in industrial recommenders},
  author={Zhu, Jie and Fan, Zhifang and Zhu, Xiaoxie and Jiang, Yuchen and Wang, Hangyu and Han, Xintian and Ding, Haoran and Wang, Xinmin and Zhao, Wenlin and Gong, Zhen and others},
  booktitle={Proceedings of the 34th ACM International Conference on Information and Knowledge Management},
  pages={6309--6316},
  year={2025}
}

@article{deepfm,
  title={DeepFM: a factorization-machine based neural network for CTR prediction},
  author={Guo, Huifeng and Tang, Ruiming and Ye, Yunming and Li, Zhenguo and He, Xiuqiang},
  journal={arXiv preprint arXiv:1703.04247},
  year={2017}
}

@article{qwen3,
  title={Qwen3 technical report},
  author={Yang, An and Li, Anfeng and Yang, Baosong and Zhang, Beichen and Hui, Binyuan and Zheng, Bo and Yu, Bowen and Gao, Chang and Huang, Chengen and Lv, Chenxu and others},
  journal={arXiv preprint arXiv:2505.09388},
  year={2025}
}

@article{deepseekv3,
  title={Deepseek-v3 technical report},
  author={Liu, Aixin and Feng, Bei and Xue, Bing and Wang, Bingxuan and Wu, Bochao and Lu, Chengda and Zhao, Chenggang and Deng, Chengqi and Zhang, Chenyu and Ruan, Chong and others},
  journal={arXiv preprint arXiv:2412.19437},
  year={2024}
}

@article{gemma2,
  title={Gemma 2: Improving open language models at a practical size},
  author={Team, Gemma and Riviere, Morgane and Pathak, Shreya and Sessa, Pier Giuseppe and Hardin, Cassidy and Bhupatiraju, Surya and Hussenot, L{\'e}onard and Mesnard, Thomas and Shahriari, Bobak and Ram{\'e}, Alexandre and others},
  journal={arXiv preprint arXiv:2408.00118},
  year={2024}
}

@article{gpt4,
  title={Gpt-4 technical report},
  author={Achiam, Josh and Adler, Steven and Agarwal, Sandhini and Ahmad, Lama and Akkaya, Ilge and Aleman, Florencia Leoni and Almeida, Diogo and Altenschmidt, Janko and Altman, Sam and Anadkat, Shyamal and others},
  journal={arXiv preprint arXiv:2303.08774},
  year={2023}
}

@article{kimik2,
  title={Kimi k2: Open agentic intelligence},
  author={Team, Kimi and Bai, Yifan and Bao, Yiping and Charles, Y and Chen, Cheng and Chen, Guanduo and Chen, Haiting and Chen, Huarong and Chen, Jiahao and Chen, Ningxin and others},
  journal={arXiv preprint arXiv:2507.20534},
  year={2025}
}

@article{glm5.0,
  title={Glm-5: from vibe coding to agentic engineering},
  author={Zeng, Aohan and Lv, Xin and Hou, Zhenyu and Du, Zhengxiao and Zheng, Qinkai and Chen, Bin and Yin, Da and Ge, Chendi and Huang, Chenghua and Xie, Chengxing and others},
  journal={arXiv preprint arXiv:2602.15763},
  year={2026}
}

@article{caafe,
  title={Large language models for automated data science: Introducing caafe for context-aware automated feature engineering},
  author={Hollmann, Noah and M{\"u}ller, Samuel and Hutter, Frank},
  journal={Advances in Neural Information Processing Systems},
  volume={36},
  pages={44753--44775},
  year={2023}
}

@article{automl-agent,
  title={Automl-agent: A multi-agent llm framework for full-pipeline automl},
  author={Trirat, Patara and Jeong, Wonyong and Hwang, Sung Ju},
  journal={arXiv preprint arXiv:2410.02958},
  year={2024}
}

@article{mle-star,
  title={Mle-star: Machine learning engineering agent via search and targeted refinement},
  author={Nam, Jaehyun and Yoon, Jinsung and Chen, Jiefeng and Shin, Jinwoo and Arik, Sercan and Pfister, Tomas},
  journal={Advances in Neural Information Processing Systems},
  volume={38},
  pages={116692--116712},
  year={2026}
}

@article{ai-scientist,
  title={The ai scientist: Towards fully automated open-ended scientific discovery},
  author={Lu, Chris and Lu, Cong and Lange, Robert Tjarko and Foerster, Jakob and Clune, Jeff and Ha, David},
  journal={arXiv preprint arXiv:2408.06292},
  year={2024}
}

@article{ai-scientistv2,
  title={The ai scientist-v2: Workshop-level automated scientific discovery via agentic tree search},
  author={Yamada, Yutaro and Lange, Robert Tjarko and Lu, Cong and Hu, Shengran and Lu, Chris and Foerster, Jakob and Clune, Jeff and Ha, David},
  journal={arXiv preprint arXiv:2504.08066},
  year={2025}
}

@article{agent-laboratory,
  title={Agent laboratory: Using llm agents as research assistants},
  author={Schmidgall, Samuel and Su, Yusheng and Wang, Ze and Sun, Ximeng and Wu, Jialian and Yu, Xiaodong and Liu, Jiang and Moor, Michael and Liu, Zicheng and Barsoum, Emad},
  journal={Findings of the Association for Computational Linguistics: EMNLP 2025},
  pages={5977--6043},
  year={2025},
  publisher={Association for Computational Linguistics}
}

@inproceedings{ab-agent,
  title={Exploring Recommender System Evaluation: A Multi-Modal LLM Agent Framework for A/B Testing},
  author={Zhang, Wenlin and Li, Xiangyang and Ge, Qiyuan and Dong, Kuicai and Jia, Pengyue and Li, Xiaopeng and Zhang, Zijian and Wang, Maolin and Wang, Yichao and Guo, Huifeng and others},
  booktitle={Proceedings of the 32nd ACM SIGKDD Conference on Knowledge Discovery and Data Mining V. 1},
  pages={2878--2889},
  year={2026}
}

@inproceedings{ab-infra,
  title={From infrastructure to culture: A/B testing challenges in large scale social networks},
  author={Xu, Ya and Chen, Nanyu and Fernandez, Addrian and Sinno, Omar and Bhasin, Anmol},
  booktitle={Proceedings of the 21th ACM SIGKDD international conference on knowledge discovery and data mining},
  pages={2227--2236},
  year={2015}
}

@inproceedings{ab-online,
  title={Online controlled experimentation at scale: an empirical survey on the current state of A/B testing},
  author={Fabijan, Aleksander and Dmitriev, Pavel and Olsson, Helena Holmstrom and Bosch, Jan},
  booktitle={2018 44th Euromicro Conference on Software Engineering and Advanced Applications (SEAA)},
  pages={68--72},
  year={2018},
  organization={IEEE}
}

@article{moment-cross,
  title={Moment\&Cross: Next-Generation Real-Time Cross-Domain CTR Prediction for Live-Streaming Recommendation at Kuaishou},
  author={Cao, Jiangxia and Wang, Shen and Li, Yue and Wang, Shenghui and Tang, Jian and Wang, Shiyao and Yang, Shuang and Liu, Zhaojie and Zhou, Guorui},
  journal={arXiv preprint arXiv:2408.05709},
  year={2024}
}

@inproceedings{autoweka,
  title={Auto-WEKA: Combined selection and hyperparameter optimization of classification algorithms},
  author={Thornton, Chris and Hutter, Frank and Hoos, Holger H and Leyton-Brown, Kevin},
  booktitle={Proceedings of the 19th ACM SIGKDD international conference on Knowledge discovery and data mining},
  pages={847--855},
  year={2013}
}

@article{ds-agent,
  title={Ds-agent: Automated data science by empowering large language models with case-based reasoning},
  author={Guo, Siyuan and Deng, Cheng and Wen, Ying and Chen, Hechang and Chang, Yi and Wang, Jun},
  journal={arXiv preprint arXiv:2402.17453},
  year={2024}
}

@inproceedings{data-interpreter,
  title={Data interpreter: An llm agent for data science},
  author={Hong, Sirui and Lin, Yizhang and Liu, Bang and Liu, Bangbang and Wu, Binhao and Zhang, Ceyao and Li, Danyang and Chen, Jiaqi and Zhang, Jiayi and Wang, Jinlin and others},
  booktitle={Findings of the Association for Computational Linguistics: ACL 2025},
  pages={19796--19821},
  year={2025}
}

@article{improve,
  title={Improve: Iterative model pipeline refinement and optimization leveraging LLM experts},
  author={Xue, Eric and Chen, Ke and Huang, Zeyi and Ji, Yuyang and Wang, Haohan},
  journal={arXiv preprint arXiv:2502.18530},
  year={2025}
}

@article{automind,
  title={Automind: Adaptive knowledgeable agent for automated data science},
  author={Ou, Yixin and Luo, Yujie and Zheng, Jingsheng and Wei, Lanning and Yu, Zhuoyun and Qiao, Shuofei and Zhang, Jintian and Zheng, Da and Mao, Yuren and Gao, Yunjun and others},
  journal={arXiv preprint arXiv:2506.10974},
  year={2025}
}

@article{mlevolve,
  title={MLEvolve: A Self-Evolving Framework for Automated Machine Learning Algorithm Discovery},
  author={Du, Shangheng and Yan, Xiangchao and Shi, Jinxin and Cao, Zongsheng and Feng, Shiyang and Liang, Zichen and Sun, Boyuan and Peng, Tianshuo and Zhou, Yifan and Li, Xin and others},
  journal={arXiv preprint arXiv:2606.06473},
  year={2026}
}

@article{rd-agent,
  title={R\&D-Agent: An LLM-Agent Framework Towards Autonomous Data Science},
  author={Yang, Xu and Yang, Xiao and Fang, Shikai and Zhang, Yifei and Wang, Jian and Xian, Bowen and Li, Qizheng and Li, Jingyuan and Xu, Minrui and Li, Yuante and others},
  journal={arXiv preprint arXiv:2505.14738},
  year={2025}
}

@article{ai-researcher,
  title={Ai-researcher: Autonomous scientific innovation},
  author={Tang, Jiabin and Xia, Lianghao and Li, Zhonghang and Huang, Chao},
  journal={Advances in Neural Information Processing Systems},
  volume={38},
  pages={9481--9520},
  year={2026}
}

@inproceedings{autoctr,
  title={Towards automated neural interaction discovery for click-through rate prediction},
  author={Song, Qingquan and Cheng, Dehua and Zhou, Hanning and Yang, Jiyan and Tian, Yuandong and Hu, Xia},
  booktitle={Proceedings of the 26th ACM SIGKDD International Conference on Knowledge Discovery \& Data Mining},
  pages={945--955},
  year={2020}
}

@inproceedings{nasrec,
  title={NASRec: weight sharing neural architecture search for recommender systems},
  author={Zhang, Tunhou and Cheng, Dehua and He, Yuchen and Chen, Zhengxing and Dai, Xiaoliang and Xiong, Liang and Yan, Feng and Li, Hai and Chen, Yiran and Wen, Wei},
  booktitle={Proceedings of the ACM Web Conference 2023},
  pages={1199--1207},
  year={2023}
}

@inproceedings{autofis,
  title={Autofis: Automatic feature interaction selection in factorization models for click-through rate prediction},
  author={Liu, Bin and Zhu, Chenxu and Li, Guilin and Zhang, Weinan and Lai, Jincai and Tang, Ruiming and He, Xiuqiang and Li, Zhenguo and Yu, Yong},
  booktitle={proceedings of the 26th ACM SIGKDD international conference on knowledge discovery \& data mining},
  pages={2636--2645},
  year={2020}
}

@article{selfvolverec,
  title={Self-EvolveRec: Self-Evolving Recommender Systems with LLM-based Directional Feedback},
  author={Kim, Sein and Park, Sangwu and Kang, Hongseok and Kim, Wonjoong and Seo, Jimin and In, Yeonjun and Yoon, Kanghoon and Park, Chanyoung},
  journal={arXiv preprint arXiv:2602.12612},
  year={2026}
}

@article{self-evolving-recsys,
  title={Self-evolving recommendation system: End-to-end autonomous model optimization with LLM agents},
  author={Wang, Haochen and Wu, Yi and Chang, Daryl and Wei, Li and Heldt, Lukasz},
  journal={arXiv preprint arXiv:2602.10226},
  year={2026}
}

@article{agenticrectune,
  title={AgenticRecTune: Multi-Agent with Self-Evolving Skillhub for Recommendation System Optimization},
  author={Wu, Xidong and Zhuan, Yue and Wei, Ruoqiao and Chen, Hangxin and Bai, Di and Liu, Jintao and Wang, Xinyi and Wang, Xue and Wang, Luoshu and Cheng, Xinwu},
  journal={arXiv preprint arXiv:2604.26969},
  year={2026}
}

@inproceedings{kuairand,
  title={Kuairand: An unbiased sequential recommendation dataset with randomly exposed videos},
  author={Gao, Chongming and Li, Shijun and Zhang, Yuan and Chen, Jiawei and Li, Biao and Lei, Wenqiang and Jiang, Peng and He, Xiangnan},
  booktitle={Proceedings of the 31st ACM international conference on information \& knowledge management},
  pages={3953--3957},
  year={2022}
}

@inproceedings{taobao,
  title={Learning tree-based deep model for recommender systems},
  author={Zhu, Han and Li, Xiang and Zhang, Pengye and Li, Guozheng and He, Jie and Li, Han and Gai, Kun},
  booktitle={Proceedings of the 24th ACM SIGKDD international conference on knowledge discovery \& data mining},
  pages={1079--1088},
  year={2018}
}

@inproceedings{mhft,
  title={Multi-granularity interest retrieval and refinement network for long-term user behavior modeling in ctr prediction},
  author={Xu, Xiang and Wang, Hao and Guo, Wei and Zhang, Luankang and Yang, Wanshan and Yu, Runlong and Liu, Yong and Lian, Defu and Chen, Enhong},
  booktitle={Proceedings of the 31st ACM SIGKDD Conference on Knowledge Discovery and Data Mining V. 1},
  pages={2745--2755},
  year={2025}
}

@article{hstu,
  title={Actions speak louder than words: Trillion-parameter sequential transducers for generative recommendations},
  author={Zhai, Jiaqi and Liao, Lucy and Liu, Xing and Wang, Yueming and Li, Rui and Cao, Xuan and Gao, Leon and Gong, Zhaojie and Gu, Fangda and He, Michael and others},
  journal={arXiv preprint arXiv:2402.17152},
  year={2024}
}

@article{mvsf,
  title={Beyond Dense Connectivity: Explicit Sparsity for Scalable Recommendation},
  author={Yu, Yantao and Qiao, Sen and Shen, Lei and Wang, Bing and Zeng, Xiaoyi},
  journal={arXiv preprint arXiv:2604.08011},
  year={2026}
}

@inproceedings{qarm,
  title={Qarm: Quantitative alignment multi-modal recommendation at kuaishou},
  author={Luo, Xinchen and Cao, Jiangxia and Sun, Tianyu and Yu, Jinkai and Huang, Rui and Yuan, Wei and Lin, Hezheng and Zheng, Yichen and Wang, Shiyao and Hu, Qigen and others},
  booktitle={Proceedings of the 34th ACM International Conference on Information and Knowledge Management},
  pages={5915--5922},
  year={2025}
}

@article{smes,
  title={SMES: Towards Scalable Multi-Task Recommendation via Expert Sparsity},
  author={Zhang, Yukun and Dong, Si and Wang, Xu and Chen, Bo and Jia, Qinglin and Wang, Shengzhe and Jiao, Jinlong and Li, Runhan and Liu, Jiaqing and Ma, Chaoyi and others},
  journal={arXiv preprint arXiv:2602.09386},
  year={2026}
}

@article{mixformer,
  title={MixFormer: Co-Scaling Up Dense and Sequence in Industrial Recommenders},
  author={Huang, Xu and Zhang, Hao and Fan, Zhifang and Huang, Yunwen and Wei, Zhuoxing and Chai, Zheng and Ni, Jinan and Zheng, Yuchao and Chen, Qiwei},
  journal={arXiv preprint arXiv:2602.14110},
  year={2026}
}

@article{sort,
  title={SORT: A Systematically Optimized Ranking Transformer for Industrial-scale Recommenders},
  author={Wang, Chunqi and Wu, Bingchao and Pang, Taotian and Wang, Jiahao and Yang, Jie and Liu, Jia and Zhang, Hao and Zhu, Hai and Shen, Lei and Wang, Shizhun and others},
  journal={arXiv preprint arXiv:2603.03988},
  year={2026}
}

@inproceedings{amazon,
  title={Justifying recommendations using distantly-labeled reviews and fine-grained aspects},
  author={Ni, Jianmo and Li, Jiacheng and McAuley, Julian},
  booktitle={Proceedings of the 2019 conference on empirical methods in natural language processing and the 9th international joint conference on natural language processing (EMNLP-IJCNLP)},
  pages={188--197},
  year={2019}
}

@article{case,
  title={CASE: Cadence-Aware Set Encoding for Large-Scale Next Basket Repurchase Recommendation},
  author={Cao, Yanan and Ranjan, Ashish and Subramaniam, Sinduja and Korpeoglu, Evren and Nag, Kaushiki and Achan, Kannan},
  journal={arXiv preprint arXiv:2604.06718},
  year={2026}
}

@article{hisac,
  title={HiSAC: Hierarchical Sparse Activation Compression for Ultra-long Sequence Modeling in Recommenders},
  author={Yuan, Kun and Bi, Junyu and Cheng, Daixuan and Wu, Changfa and Xiao, Shuwen and Cao, Binbin and Wu, Jian and Jiang, Yuning},
  journal={arXiv preprint arXiv:2602.21009},
  year={2026}
}

@article{laser,
  title={LASER: An Efficient Target-Aware Segmented Attention Framework for End-to-End Long Sequence Modeling},
  author={Lin, Tianhe and Xiong, Ziwei and Ou, Baoyuan and Qin, Yingjie and Xu, Lai and Zhong, Xiaocheng and Hu, Yao and Wang, Zhiyong and Zhou, Tao and Xu, Yubin and others},
  journal={arXiv preprint arXiv:2602.11562},
  year={2026}
}

@article{zenith,
  title={Zenith: Scaling up Ranking Models for Billion-scale Livestreaming Recommendation},
  author={Zhang, Ruifeng and Huang, Zexi and Wang, Zikai and Sun, Ke and Zheng, Bohang and Jiang, Yuchen and Chen, Zhe and Ouyang, Zhen and Xie, Huimin and Shen, Phil and others},
  journal={arXiv preprint arXiv:2601.21285},
  year={2026}
}

@article{wukong,
  title={Wukong: Towards a scaling law for large-scale recommendation},
  author={Zhang, Buyun and Luo, Liang and Chen, Yuxin and Nie, Jade and Liu, Xi and Guo, Daifeng and Zhao, Yanli and Li, Shen and Hao, Yuchen and Yao, Yantao and others},
  journal={arXiv preprint arXiv:2403.02545},
  year={2024}
}

@article{movielens,
  title={The movielens datasets: History and context},
  author={Harper, F Maxwell and Konstan, Joseph A},
  journal={Acm transactions on interactive intelligent systems (tiis)},
  volume={5},
  number={4},
  pages={1--19},
  year={2015},
  publisher={Acm New York, NY, USA}
}

@article{rankup,
  title={RankUp: Towards High-rank Representations for Large Scale Advertising Recommender Systems},
  author={Chen, Jin and Zhang, Shangyu and Hu, Bin and Zhou, Chao and Pan, Junwei and Xue, Gengsheng and Ning, Wentao and Weng, Gengyu and Zheng, Wang and Liu, Shaohua and others},
  journal={arXiv preprint arXiv:2604.17878},
  year={2026}
}

\newpage

\appendix

\section{Author List}

\noindent\textbf{Core Contributors}\quad
Changxin Lao, Fei Pan, Guozhuang Ma, Han Li, Huihuang Lin, Jijun Shi, Kangzhi Zhao, Kun Gai, Mo Zhou, Qinqin Zhou, Quan Chen, Ruochen Yang, Shifu Bie, Shijie Yi, Shuang Yang, Shuo Yang, Wenhao Li, Wentao Xie, Xiao Lv, Xuming Wang, Yijun Wang, Yiming Chen, Yusheng Huang, Zhongyuan Wang, Zibo Zhao, Zijie Zhuang

\vspace{0.8em}
\noindent\textbf{Contributors}\quad
Baoning Xia, Chao Liu, Chaoyi Ma, Chubo He, Dawei Cong, Feng Jiang, Gang Wang, Guilin Xia, Hanwen Xu, Jiahong Xie, Jiahui Qiao, Jian Liang, Jiangfan Yue, Jing Wang, Jinghan Yang, Jinghui Jia, Kan Qin, Lei Wang, Ming Li, Peilin Song, Pengbo Xu, Qiang Luo, Ruiming Tang, Shiyang Liu, Shuxian Jin, Tao Wang*, Tao Zhang, Xiang Gao, Xianghan Li, Yingsong Luo, Yiwen Ning, Yongcheng Liu, Yueyang Liu, Yuan Guo, Zhaojie Liu, Zhenkai Cui

\vspace{1.2em}
\noindent All the authors listed alphabetically by first name.\qquad $^*$individuals who have departed from our team.

\section{Model Research Experiment}
\label{sec:detailed_experiment}
\subsection{Reproduction Experiment}

\begin{table}[h]
\centering
\caption{Performance comparison of reproduced models across multiple methods. The baseline results are \textbf{boldfaced}, the optimal results are \underline{underlined}.}
\label{tab:rankmixer}
\begin{tabular}{lcccccccc}
\toprule
\multirow{2.5}{*}{\textbf{Model}} & \multicolumn{2}{c}{\textbf{KuaiRand}} & \multicolumn{2}{c}{\textbf{Taobao}} & \multicolumn{2}{c}{\textbf{Amazon}} & \multicolumn{2}{c}{\textbf{ML-1M}} \\
\cmidrule(lr){2-3} \cmidrule(lr){4-5} \cmidrule(lr){6-7} \cmidrule(lr){8-9}
 & AUC & \textit{Imprv.}$\uparrow$ & AUC & \textit{Imprv.}$\uparrow$ & AUC & \textit{Imprv.}$\uparrow$ & AUC & \textit{Imprv.}$\uparrow$ \\

\midrule
\textbf{RankMixer} & \textbf{0.6860} & - & \textbf{0.5647} & - & \textbf{0.6671} & - & \textbf{0.7935} & - \\
\midrule
CASE & 0.6643 & -0.0217 & 0.6090 & +0.0443 & \underline{0.6913} & \underline{+0.0242} & 0.7947 & +0.0012 \\
HiSAC & 0.6860 & +0.0000 & 0.6099 & +0.0452 & 0.6671 & +0.0000 & 0.7935 & +0.0000 \\
SORT & 0.6523 & -0.0337 & \underline{0.6189} & \underline{+0.0542} & 0.6770 & +0.0099 & \underline{0.8020} & \underline{+0.0085} \\
LASER & 0.6860 & +0.0000 & 0.5881 & +0.0234 & 0.6671 & +0.0000 & 0.7935 & +0.0000 \\
Zenith & 0.6895 & +0.0035 & 0.5821 & +0.0174 & 0.6732 & +0.0061 & 0.7864 & -0.0071 \\
RankUp & 0.6742 & -0.0118 & 0.5780 & +0.0133 & 0.6808 & +0.0137 & 0.7948 & +0.0013 \\
Wukong & \underline{0.6897} & \underline{+0.0037} & 0.5685 & +0.0038 & 0.6588 & -0.0083 & 0.7881 & -0.0054 \\
QARM & 0.6879 & +0.0019 & 0.5712 & +0.0065 & 0.6690 & +0.0019 & 0.7936 & +0.0001 \\

\bottomrule
\end{tabular}
\end{table}

A prerequisite for autonomous model exploration is the ability to faithfully reproduce recent published recommendation methods. We therefore evaluate AgentX from this foundational angle, assessing its capability to translate papers into executable and reproducible engineering artifacts. 
Specifically, AgentX is required to read each target paper, decompose its proposed mechanism into reusable components, implement these components on top of a unified backbone architecture, and run training and evaluation under a unified protocol.
We use RankMixer~\cite{rankmixer} as the base model and evaluate on four public recommendation datasets, KuaiRand~\cite{kuairand}, Taobao~\cite{taobao}, Amazon~\cite{amazon} and ML-1M~\cite{movielens}. 
The reproduced methods include CASE~\cite{case}, HiSAC~\cite{hisac}, SORT~\cite{sort}, LASER~\cite{laser}, Zenith~\cite{zenith}, RankUp~\cite{rankup}, Wukong~\cite{wukong} and QARM~\cite{qarm}, among others.
We use AUC as the main evaluation metric and demonstrate the results in Table \ref{tab:rankmixer}.

The results show that AgentX completes the full reproduction loop for every (method, dataset) pair, producing 32 trainable variants on top of the same RankMixer backbone across heterogeneous mechanisms. This indicates that AgentX can faithfully translate diverse paper-level designs into runnable artifacts under a unified protocol.

Moreover, no single method dominates across all four datasets.
The fact that AgentX surfaces these cross-dataset discrepancies, rather than reporting uniformly positive numbers, suggests that its reproductions reflect genuine method behavior and provide trustworthy building blocks for the subsequent module-exploration loop.

\subsection{Module Exploration Experiment}

\begin{table}[h]
\centering
\caption{Performance comparison of multi-module combination results on public datasets based on RankMixer. The baseline results are \textbf{boldfaced}, the optimal results for different numbers of multi-module combination are \underline{underlined}, and the results of the overall optimal combination are highlighted in \textit{blue}.}
\label{tab:module_fusionm}
\begin{tabular}{lccccc}
\toprule
\multirow{2.5}{*}{\textbf{Variants}} & \multicolumn{2}{c}{\textbf{KuaiRand}} & \multicolumn{2}{c}{\textbf{Taobao}} & \multirow{2.5}{*}{\textbf{Avg. \textit{Imprv.}$\uparrow$}} \\
\cmidrule(lr){2-3} \cmidrule(lr){4-5} 
 & AUC & \textit{Imprv.}$\uparrow$ & AUC & \textit{Imprv.}$\uparrow$ & \\

\midrule

\textbf{RankMixer (Baseline)} & \textbf{0.6860} & \textbf{-} & \textbf{0.5647} & \textbf{-} & \textbf{-} \\

\midrule
\multicolumn{6}{l}{\textit{Single module}} \\
\quad + MHFT & 0.6871 & +0.0011 & \underline{0.6151} & \underline{+0.0504} & \underline{+0.0258} \\
\quad + HSTU & 0.6908 & +0.0048 & 0.5765 & +0.0118 & +0.0083 \\
\quad + MVSF & \underline{0.6924} & \underline{+0.0064} & 0.5748 & +0.0101 & +0.0083 \\
\quad + VQ\_Code & 0.6876 & +0.0016 & 0.5720 & +0.0073 & +0.0045 \\
\quad + SMES & 0.6886 & +0.0026 & 0.5666 & +0.0019 & +0.0023 \\
\quad + SORT & 0.6923 & +0.0063 & 0.5771 & +0.0124 & +0.0094 \\
\quad + MixFormer & 0.6838 & -0.0022 & 0.5722 & +0.0075 & +0.0027 \\

\midrule
\multicolumn{6}{l}{\textit{Two modules}} \\
\quad + MHFT + HSTU & \underline{0.6895} & \underline{+0.0035} & \underline{0.6017} & \underline{+0.0370} & \underline{+0.0203} \\
\quad + MHFT + SMES & 0.6886 & +0.0026 & 0.6006 & +0.0359 & +0.0193 \\
\quad + MHFT + VQ\_Code & 0.6860 & +0.0000 & 0.5989 & +0.0342 & +0.0171 \\
\quad + MHFT + MixFormer & 0.6826 & -0.0034 & 0.5963 & +0.0316 & +0.0141 \\
\quad + SORT + MixFormer & 0.6823 & -0.0037 & 0.5783 & +0.0136 & +0.0050 \\
\quad + SORT + MVSF & 0.6856 & -0.0004 & 0.5670 & +0.0023 & +0.0010 \\

\midrule
\multicolumn{6}{l}{\textit{Three modules}} \\
\quad + MHFT + HSTU + SMES & 0.6899 & +0.0039 & 0.6090 & +0.0443 & +0.0241 \\
\quad + MHFT + HSTU + MVSF & 0.6895 & +0.0035 & 0.6017 & +0.0370 & +0.0203 \\
\rowcolor{blue!8}
\textbf{\quad + MHFT + VQ\_Code + MVSF} & \underline{\textbf{0.6900}} & \underline{\textbf{+0.0040}} & \underline{\textbf{0.6134}} & \underline{\textbf{+0.0487}} & \underline{\textbf{+0.0264}} \\

\bottomrule
\end{tabular}
\end{table}

To evaluate whether AgentX can autonomously discover and implement effective improvements for recommendation models based on existing studies and recent research knowledge, we conduct an offline self-exploration experiment. Specifically, we also use RankMixer as the base model and evaluate on two public recommendation datasets, KuaiRand and Taobao. AgentX is instructed to automatically retrieve and understand recent related works, propose compatible model modules, implement the corresponding code changes, run training and evaluation, and select promising candidates. 
The modules explored include:
\begin{itemize}
\item MHFT (Multi-Head Fourier Transformer)~\cite{mhft}: refines multi-granularity user interests by modeling sequential patterns with Fourier-based transformations.
\item HSTU (Hierarchical Sequential Transduction Units)~\cite{hstu}: enables scalable long-history behavior modeling through efficient sequential transduction.
\item MVSF (Multi-View Sparse Filtering)~\cite{mvsf}: improves recommendation scalability by selecting informative sparse signals from multiple feature views.
\item VQ\_Code~\cite{qarm}: transforms multi-modal item representations into discrete semantic codes for downstream recommendation models.
\item SMES (Scalable Multi-Task via Expert Sparsity)~\cite{smes}: uses sparse expert routing to support efficient and scalable multi-task recommendation learning.
\item SORT (Systematically Optimized Ranking Transformer)~\cite{sort}: enhances industrial ranking performance through systematic optimization of Transformer-based recommendation architecture.
\item MixFormer~\cite{mixformer}: jointly captures dense feature interactions and sequential user behaviors within a unified recommendation backbone.
\end{itemize}
We use AUC as the main evaluation metric and report performance comparison of multi-module combination results in Table \ref{tab:module_fusionm}.

The results show that AgentX can extract transferable model designs from external research knowledge and convert them into executable and evaluable engineering implementations. In the single-module setting, multiple candidates outperform the RankMixer baseline. For example, MVSF and SORT improve AUC on KuaiRand by +0.0064 and +0.0063, respectively, while MHFT brings a +0.0504 improvement on Taobao. These results suggest that the candidates generated by AgentX are not simple random trials, but meaningful model modules with practical improvement potential.

Furthermore, AgentX is also capable of exploring the combinations of different modules, and covering the complementarity among different modeling mechanisms by combining multiple modules.
In the multi-module settings, several combinations achieve positive gains. Among them, MHFT + VQ-Code + MVSF obtains AUC scores of 0.6900 and 0.6134 on KuaiRand and Taobao, respectively, with an average improvement of +0.0264, yielding the best overall performance. 
This indicates that AgentX can not only identify effective individual modules, but also discover stronger model variants through combinatorial exploration.

Overall, this experiment validates the effectiveness of AgentX in offline recommendation model exploration. AgentX can autonomously propose candidate improvements based on recent research knowledge, translate paper-level modules into runnable engineering implementations, and select valuable modules and module combinations through automatic experimentation.

\subsection{Industrial Scenario Experiment}

\begin{table}[h]
\centering
\caption{Performance comparison of multi-module combination results on industrial scenario dataset. The best result is highlighted in \textit{blue}.}
\label{tab:rtb}
\begin{tabular}{lc}
\toprule

\textbf{Variants} & \textbf{AUC\_\textit{Imprv.}$\uparrow$} \\

\midrule

\textbf{Rankmixer (Baseline)} & - \\

\midrule

\rowcolor{blue!8}
\quad \textbf{+ MHFT} & \textbf{+0.0151} \\
\quad + MVSF & +0.0057 \\
\quad + VQ\_Code & +0.0028 \\
\quad + MHFT + SMES & +0.0092 \\
\quad + MHFT + HSTU & +0.0091 \\
\quad + MHFT + VQ\_Code & +0.0086 \\
\quad + MHFT + HSTU + MVSF & +0.0098 \\

\bottomrule
\end{tabular}
\end{table}

To further verify whether the modules and combinations identified on public datasets remain effective in real industrial recommendation, we transfer a subset of the better-performing variants above to a production-scale dataset from an advertising delivery and user growth scenario, which contains million-level users and billion-level interactions.
Each selected variant is plugged into the same RankMixer backbone and undergoes training and evaluation under the production protocol, so that the comparison reflects how each design behaves once it is exposed to large-scale, long-tailed industrial data rather than curated public splits. 
We report the relative AUC improvement over the RankMixer baseline in Table~\ref{tab:rtb}.

The industrial results are broadly consistent with the offline observations on public datasets. The single-module MHFT and the three-module MHFT + HSTU + MVSF, which already rank among the strongest variants on KuaiRand and Taobao, also lead the industrial leaderboard with $+0.0151$ and $+0.0098$ AUC improvement, respectively. 
This suggests that the modules and combinations surfaced by AgentX carry a genuinely transferable component, rather than merely overfitting to the statistics of any single public benchmark.

At the same time, the ranking is not strictly preserved: some combinations that appeared promising offline shift in relative order under industrial data, suggesting that public-dataset evidence and industrial behavior are correlated but not equivalent. This implies that combinatorial exploration still needs to be re-validated on production data, and at the same time demonstrates that AgentX is already capable of conducting effective module exploration directly in industrial recommendation scenarios.

\section{Prompt Formats}
\subsection{Brainstorm Prompt}

\makeatletter
\@ifundefined{mycase}{%
  \newtcolorbox{mycase}[1]{%
    breakable,
    colback=gray!5, colframe=black!60, boxrule=0.5pt, arc=2pt,
    left=6pt, right=6pt, top=6pt, bottom=6pt,
    title={#1},
    fonttitle=\bfseries\small, 
    fontupper=\tiny,
    coltitle=white, colbacktitle=black!70,
    before skip=8pt, after skip=8pt,
    before upper={\raggedright\sloppy}%
  }%
}{}
\makeatother

\begin{mycase}{Prompt: Candidate Idea Generation Agent}
\small
\textbf{Role} \\
\textit{An autonomous sub-agent invoked by an upstream proposer agent. It generates candidate experimental directions for a recommendation system, but does not perform validation, final selection, or implementation handoff.}

\vspace{0.5em}
\textbf{Task} \\
Given a question brief and an input brief, produce a set of candidate experiment directions grounded in historical context, source code, prior launch-review materials, and read-only code knowledge bases. Optionally request sampled data analysis as supporting evidence. Write results to \texttt{<ROUND\_DIR>/candidate\_ideas.md} inside the proposal batch directory.

\vspace{0.5em}
\textbf{Inputs} \\
- question brief and input brief from the upstream agent\\
- prior experiment results and exploration logs\\
- launch-review archive and read-only code-knowledge wikis\\
- business-knowledge definitions: pipeline stages, metrics, segments, strategy levers, model predictions\\
- relevant source code in the recommendation service

\vspace{0.5em}
\textbf{Workflow} \\
1. Read context and anchor every candidate to the round's first objective.\\
2. Map the brainstorming space across four dimensions: pipeline stage, business metrics, user/content segments, and strategy levers.\\
3. Scan team-wide historical experiments for parameter conflicts and reusable conclusions.\\
4. Retrieve launch-review documents and query code wikis via progressive disclosure.\\
5. Cross-check model-prediction signals against business-knowledge definitions; optionally invoke a data-analysis sub-agent for supporting evidence.\\
6. Generate candidates and write the artifact.

\vspace{0.5em}
\textbf{Constraints} \\
- Cap candidates at five; do not duplicate rejected or machine-passed directions.\\
- Do not fabricate attribute semantics, parameter names, or formula inputs; mark unknowns.\\
- Stay within documented agent capabilities.

\vspace{0.5em}
\textbf{Output Format} \\
One markdown file written via the file-writing tool. The body is open-ended markdown; the only structural expectation is a set of fixed top-level sections so the downstream validator can locate the key conclusions:

\begin{verbatim}
---
input_summary: objective, domain, metrics, constraints, scope
user_constraint_mapping:
business_knowledge_mapping:
knowledge_sources:
data_evidence:
candidate_list:
    - candidate <i>:
open_questions:
historical_rejections:
---
\end{verbatim}

\end{mycase}

\begin{mycase}{Prompt: Candidate Validation Agent}
\small
\textbf{Role} \\
\textit{An autonomous sub-agent invoked by an upstream proposer agent. It reviews candidate experimental directions, assigns per-candidate status, and emits a round-level verdict. It does not generate new candidates or write implementation plans.}

\vspace{0.5em}
\textbf{Task} \\
Audit each candidate in the round's candidate-ideas artifact against objective alignment, business semantics, user constraints, model-prediction semantics, capability boundaries, historical overlap, and AB-parameter feasibility. Produce a round-level verdict in \{PASS, REVISE, BLOCK\} and a per-candidate machine status in \{PASS\_READY, PASS\_PROBE, BACKLOG, REJECTED, BLOCKED\}.

\vspace{0.5em}
\textbf{Inputs} \\
- the round's candidate-ideas artifact and proposal contract\\
- agent capability reference and prior batch shortlist\\
- launch-review archive (read-only)\\
- business-knowledge definitions: pipeline stages, metrics, segments, strategy levers, model predictions\\
- AB-parameter validation tool

\vspace{0.5em}
\textbf{Workflow} \\
1. Extract each candidate's hypothesis, positioning, readiness tier, evidence, and affected parameters or code.\\
2. Audit first-objective alignment and business semantics as the highest-priority gates.\\
3. Check capability boundaries, user-constraint compliance, and model-prediction semantics; classify signals as matched, code-verified, or unresolved.\\
4. Retrieve relevant launch-review documents to assess overlap and reusable evidence.\\
5. Validate every involved AB parameter via the provided tool.\\
6. Map readiness tier to a per-candidate machine status under strict admission rules.\\
7. Emit the round-level verdict, record excluded candidates in the unsupported-ideas artifact, and write the validation summary.

\vspace{0.5em}
\textbf{Constraints} \\
- Candidates with unresolved core signals, or outside documented agent capabilities, cannot be marked PASS\_READY.\\
- Only PASS\_READY may enter the materialization shortlist; cap PASS\_READY at two per round.\\
- Every excluded, rejected, blocked, or unselected PASS\_READY candidate must be appended to the unsupported-ideas artifact before the validation summary is written.\\
- Machine review must not read or wait for human-review records.

\vspace{0.5em}
\textbf{Output Format} \\
Two markdown files written via the file-writing tool, with fixed sections so downstream agents can parse the conclusions:

\begin{verbatim}
---
validated_idea.md:
    verdict: PASS | REVISE | BLOCK
    knowledge_sources:
    first_objective_alignment_audit:
    business_semantics_audit:
    user_constraint_compliance_audit:
    model_prediction_semantics_audit:
    readiness_audit:
    selected_directions:
    unselected_directions:
    implementation_prerequisites:

machine_review_summary.md:
    candidate_index:
        - candidate <i>:
---
\end{verbatim}

\end{mycase}

\subsection{Developing Prompt}

\begin{mycase}{Prompt: Code Implementation Agent}
\small
\textbf{Role} \\
\textit{An autonomous sub-agent invoked when an experiment proposal is classified as a code-change experiment. It executes the full coding workflow against an approved implementation plan and returns the resulting code-change metadata to the upstream proposer.}

\vspace{0.5em}
\textbf{Task} \\
Take an approved implementation plan as input and deliver a verified merge request. The agent isolates a clean feature branch from a debug branch, performs dryrun verification under both branches, and writes back the resulting artifacts to the experiment workspace.

\vspace{0.5em}
\textbf{Inputs} \\
- approved implementation plan for the experiment\\
- recommendation-service source repository\\
- coding skill reference and proposal contract\\
- static-precheck, code-review, and pipeline tools\\
- dryrun and merge-request sub-agent

\vspace{0.5em}
\textbf{Workflow} \\
1. Sync the repository, read all files to be modified together with neighboring references, and abort early if the target architecture is unfamiliar.\\
2. Create or reset the feature branch from the target branch, then iterate: write functional code, run static precheck, perform dual self-check, run local code review, commit and push, run static-check pipelines, and pass a confirmation gate.\\
3. Create or reset the debug branch from the feature branch, then iterate: add force-enable switches and always-on logs only, self-check, run an incremental review against the feature branch, run local validation, and push.\\
4. Invoke the dryrun and merge-request sub-agent to run debug dryrun, log verification, clean dryrun, and merge-request creation; route by terminal status and locate root cause before retrying.\\
5. Write code-change metadata (modified files, dryrun links, merge-request link, retry counts) back to the experiment workspace.

\vspace{0.5em}
\textbf{Constraints} \\
- Every new feature must be off by default; no production-behavior change at merge time.\\
- Clean code lives only on the feature branch; force-enable switches and logs only on the debug branch. Business-logic fixes are forbidden on the debug branch.\\
- Static precheck is a hard gate before self-check, commit, or dryrun.\\
- Commits must be incremental; an empty diff at the confirmation gate is blocking.\\
- On dryrun failure, classify root cause (clean-code defect / debug-wiring defect / infra) before deciding which loop to re-enter.

\vspace{0.5em}
\textbf{Output Format} \\
Code-change metadata appended to the experiment workspace via the file-writing tool, with fixed fields so downstream agents can parse the result:

\begin{verbatim}
---
work_in_progress.md (coding section):
    modified_files:
    debug:
    merge_request:
    retry_counts:
---
\end{verbatim}

\end{mycase}

\subsection{Evaluation Prompt}

\begin{mycase}{Prompt: Experiment Launch Agent}
\small
\textbf{Role} \\
\textit{An autonomous sub-agent invoked after an experiment proposal and its implementation plan are approved. It records the proposal summary, configures the AB-experiment parameters through a write API, updates the workspace state, and releases the experiment lock.}

\vspace{0.5em}
\textbf{Task} \\
Turn an approved implementation plan into a running AB experiment. The agent requests the AB-experiment identifier from a human-in-the-loop step, fetches the experiment configuration, scans for free buckets, and submits parameter changes under the platform's gray-rollout rules.

\vspace{0.5em}
\textbf{Inputs} \\
- approved implementation plan and proposal contract\\
- workspace state and experiment registry\\
- AB-platform information and parameter-write tools\\
- recommendation-service repository for parameter lookup

\vspace{0.5em}
\textbf{Workflow} \\
1. Append the proposal summary (objective, type, modified files, dryrun and merge-request links, AB configuration, expected impact) to the experiment workspace.\\
2. Pause and request the AB-experiment name and world from the user; do not proceed until provided.\\
3. Fetch experiment details: identifier, groups, traffic shares, and existing parameter values; generate the canonical platform page links.\\
4. Scan all experiment groups for non-default parameter differences and select a truly free bucket; refuse to overlay on an occupied bucket.\\
5. For each parameter in the implementation plan, classify as new or reused via the lookup tool, then submit additions or updates in batched form per platform constraints, with explicit gray configuration.\\
6. Handle the verification response, wait for the required gray duration, and trigger gray rollout only on explicit user authorization.\\
7. Patch the workspace state with the AB metadata, append the launch summary, and release the experiment lock.

\vspace{0.5em}
\textbf{Constraints} \\
- Default value of any newly added parameter must equal the control group's value; only experiment buckets receive explicit values.\\
- The bucket scan must be re-run at every launch; prior selections cannot be reused without verification.\\
- AB-platform page links must be built from the numeric experiment identifier via the provided tool, not hand-written from the name.\\
- The agent must not auto-create AB experiments; that step requires human action.\\
- Gray rollout must respect minimum gray duration and must not trigger automatically.

\vspace{0.5em}
\textbf{Output Format} \\
Workspace state patched via the file-writing tool, with fixed fields so downstream monitoring agents can parse the experiment record:

\begin{verbatim}
---
experiments.json (per experiment entry):
    ab_experiment_id
    ab_experiment_url
    ab_experiment_group_url
    ab_experiment_analysis_url
    ab_experiment_name
    ab_world
    ab_bucket
    status: ab_running
    deployed_at

work_in_progress.md (launch section):
    experiment_summary
    parameter_submission_result
    verify_ids
    gray_rollout_status
    lock_released
---
\end{verbatim}

\end{mycase}

\begin{mycase}{Prompt: AB-Test Monitoring Agent}
\small
\textbf{Role} \\
\textit{An autonomous sub-agent invoked after an experiment is launched. It monitors realtime and daily AB metrics, checks whether the minimum experiment duration has been reached, and signals readiness for the downstream decision agent.}

\vspace{0.5em}
\textbf{Task} \\
Given a running AB experiment, identify its platform and split type, retrieve the appropriate metric universe, pull realtime and daily metrics, judge significance and guardrails under the right analysis method, and decide whether to wait, advance to decision, or alert on regression.

\vspace{0.5em}
\textbf{Inputs} \\
- workspace experiment registry with AB identifier and link\\
- AB-platform metric-fetching and parameter-lookup tools\\
- platform configuration: minimum and maximum experiment days, primary and guardrail metrics with thresholds\\
- optional user-supplied raw data text when the API path is unavailable

\vspace{0.5em}
\textbf{Workflow} \\
1. Load the metric-fetching tool and read the platform configuration.\\
2. Determine the data source: structured experiment reference, or user-pasted text for direct extraction.\\
3. Confirm the experiment platform and the split type via the registry, the parameter lookup tool, world-name inference, or by asking the user.\\
4. Fetch the available metric universe and identify primary and guardrail metric identifiers; switch templates when split-type incompatibility produces unknown results.\\
5. Pull realtime metrics for health-check only; do not use them for significance judgments.\\
6. Pull daily metrics up to the last fully processed day, preferring the bias-corrected analysis method and falling back to plain group comparison when baseline data is unavailable.\\
7. Handle data-status codes for missing baselines, missing dates, or unsubscribed metrics.\\
8. Compare days elapsed against the configured minimum and maximum durations and produce the next-step recommendation.

\vspace{0.5em}
\textbf{Constraints} \\
- Daily metrics are the gold standard for significance; realtime colors do not imply significance.\\
- Same-day data is excluded; daily-metric end time is always the previous day.\\
- For dual-platform experiments, any single-side guardrail violation blocks a keep decision.\\
- The agent must not modify experiment parameters or trigger rollouts; observation only.

\vspace{0.5em}
\textbf{Output Format} \\
A monitoring report delivered to the user with fixed sections so downstream decision steps can parse the conclusions:

\begin{verbatim}
---
abtest_report:
    platform: kuaishou | kuaishou nebula | dual
    split_type: did | uid 
    days_elapsed
    min_required_days
    max_allowed_days
    realtime_health:
    daily_metrics:
    next_step:
    notes:
---
\end{verbatim}

\end{mycase}

\end{document}